\documentclass[%
 reprint,
 superscriptaddress,
 amsmath,amssymb,
 aps,
 multiline,
 prl,
]{revtex4-2}
\usepackage{graphicx}
\usepackage{xcolor}
\usepackage{amsmath}
\usepackage{amssymb}
\usepackage{multirow}
\usepackage{rotating}
\usepackage{float}
\usepackage[ruled,vlined]{algorithm2e}
\usepackage{algorithmic}
\usepackage{subcaption}

\renewcommand{\selectlanguage}[1]{}

\begin{document}

\title{Reinforcement Learning Techniques for the Optimization of Target Polarization in Nuclear Physics Scattering Experiments}

\date{\today}

\author{Armen Kasparian}
\affiliation{Data Science Department, Thomas Jefferson National Accelerator Facility, Newport News, VA 23606 USA}
\author{Torri Jeske}
\affiliation{EPSCI, Thomas Jefferson National Accelerator Facility, Newport News, VA 23606 USA}
\author{Monibor Rahman}
\affiliation{Data Science Department, Thomas Jefferson National Accelerator Facility, Newport News, VA 23606 USA}
\author{Chris Keith}
\affiliation{Target Group, Thomas Jefferson National Accelerator Facility, Newport News, VA 23606 USA}
\author{James Maxwell}
\affiliation{Target Group, Thomas Jefferson National Accelerator Facility, Newport News, VA 23606 USA}
\author{Thomas Britton}
\affiliation{EPSCI, Thomas Jefferson National Accelerator Facility, Newport News, VA 23606 USA}
\author{Malachi Schram}
\affiliation{Advanced Computing, Mathematics, and Data (ACMD) Division, Pacific Northwest National Lab, Richland, WA 99354 USA} 
\author{David Lawrence}
\affiliation{EPSCI, Thomas Jefferson National Accelerator Facility, Newport News, VA 23606 USA}

\begin{abstract}
The operation of dynamically polarized targets in nuclear physics experiments relies on continuous tuning of the microwave frequency to compensate for radiation damage and evolving material properties, a task that is traditionally performed through manual trial-and-error by expert operators. This work presents a data-driven control framework that combines surrogate modeling with reinforcement learning to optimize the target polarization. Using operational data from the APOLLO cryogenic target system, we train and evaluate multilayer perceptron and Gaussian process regression models to predict polarization as a function of microwave frequency, beam current, and accumulated radiation dose. We show that Gaussian process-based models provide calibrated uncertainty estimates and reliably identify regions outside the training distribution, while MLPs exhibit limited sensitivity to distributional shift. To enable learning and control across multiple target samples, we introduce a Gaussian process approximation and embed the surrogate model within a standardized simulation environment. A reinforcement learning agent is trained using a lower-confidence-bound reward formulation that balances performance maximization against uncertainty. We are able to show an almost 2x improvement on the operators actions utilizing our RL agent.
\end{abstract}

\maketitle

\section{1 Introduction}
Nuclear physics facilities increasingly depend on complex, tightly coupled experimental subsystems that must operate continuously and reliably over extended run periods. In experimental halls at the Thomas Jefferson National Accelerator Facility (Jefferson Lab), maintaining stable performance requires frequent low-level adjustments to compensate for environmental drift, radiation-induced degradation, and evolving material properties across these subsystems. These adjustments are traditionally performed manually by expert operators using experience-driven heuristics and continuous monitoring. While effective, this approach places significant cognitive and operational demands on a small number of specialists and becomes increasingly difficult to sustain as experiments grow in duration and complexity. This has motivated growing interest in control systems that can assist human operators by adapting to changing conditions in a consistent, reliable, and data-driven manner.

Despite significant progress in artificial intelligence and machine learning for control and optimization in accelerator facilities, the application of ML techniques for controlling nuclear physics experimental subsystems remains relatively limited. Machine learning has seen widespread adoption in offline data analysis tasks, including event reconstruction, classification, and simulation, \cite{RevModPhys.94.031003}, with comparatively little attention has been given to online or near real-time control of experimental subsystems. This gap is driven in part by the stringent reliability, safety, and interpretability requirements of scientific facilities, as well as the prevalence of legacy control systems. As a result, experimental halls represent a largely unexplored application space for applied machine learning, with substantial potential impact if these constraints can be properly addressed. 

Jefferson Lab's spin-polarized solid targets present one such attractive candidate for the application of machine learning methods. These targets provide large ensembles of protons or deuterons with their spins aligned, or \textit{polarized}, in a particular direction; increasing the degree of polarization improves the statistical significance of spin-dependent scattering experiments. Using ML techniques to replace or augment the effort of human operators who are charged with optimizing the polarization would increase the impact of these experiments while reducing time, effort, and cost.

This paper is outlined as follows. In Section 2, we provide an overview of polarized target systems used at Jefferson Lab. Section 3 details previous work of ML for scientific controls. Section 4 gives a background of the data processing, models and algorithms used. Section 5 provides an overview of the trained and calibrated surrogate models used in the analysis as well as the simulated environment. Section 6 demonstrates the results and performance analysis. Conclusions and future work are outlined in Section 7.   

\section{2 Polarized Targets at Jefferson Lab}
\label{sec:polarized_targets}
Jefferson Lab operates the Continuous Electron Beam Accelerator Facility (CEBAF), a 12\,GeV electron accelerator for performing basic nuclear and particle research in four experimental halls, A, B, C, and D.  In these halls, high energy beams of electrons or photons are scattered from a variety of nuclear targets to probe the dynamics of nucleons and nuclei. 
\textit{Polarized targets}, in which the nuclear spins of the material have a net alignment along a chosen direction, enable precision measurements of the spin and electromagnetic structure of nuclear matter~\cite{GOERTZ2002403}. Polarized bulk solid target material is produced at JLab via Dynamic Nuclear Polarization (DNP).  

In DNP, low temperatures and high magnetic fields are used to create very high electron polarization in paramagnetic radicals within the material. 
Microwave radiation near the electrons' spin resonance (ESR) frequency is then applied to drive spin-flip transitions that transfer the electron polarization to nearby nuclei.  The nuclear polarization can be made parallel or anti-parallel to the magnetic field by adjusting the microwave frequency slightly below or above the ESR frequency, with the optimum frequencies largely determined by the width of the ESR line.

The most commonly utilized DNP materials at JLab are solid ammonia (NH$_3$) and deuterated ammonia (ND$_3$), which can provide proton (deuteron) polarizations in excess of 95\% (50\%) at operational conditions of 5\,T and 1\,K.  The paramagnetic radicals in these materials are produced \textit{in situ} with ionizing radiation; an initial dose of approximately 10$^{17}$e$^-$/cm$^2$ at a temperature of 80\,K yields an optimum concentration for DNP.  

As additional radiation from the experimental electron beam produces an overabundance of free radicals, the width of the ESR line increases, and the target polarization decreases in a roughly exponential manner.  The rate of this polarization loss can be reduced by regular adjustment of the microwave frequency. Failure to maintain the optimum frequency means the polarization will drop more rapidly. 

\begin{figure}
    \centering
    \includegraphics[width=1.0\linewidth]{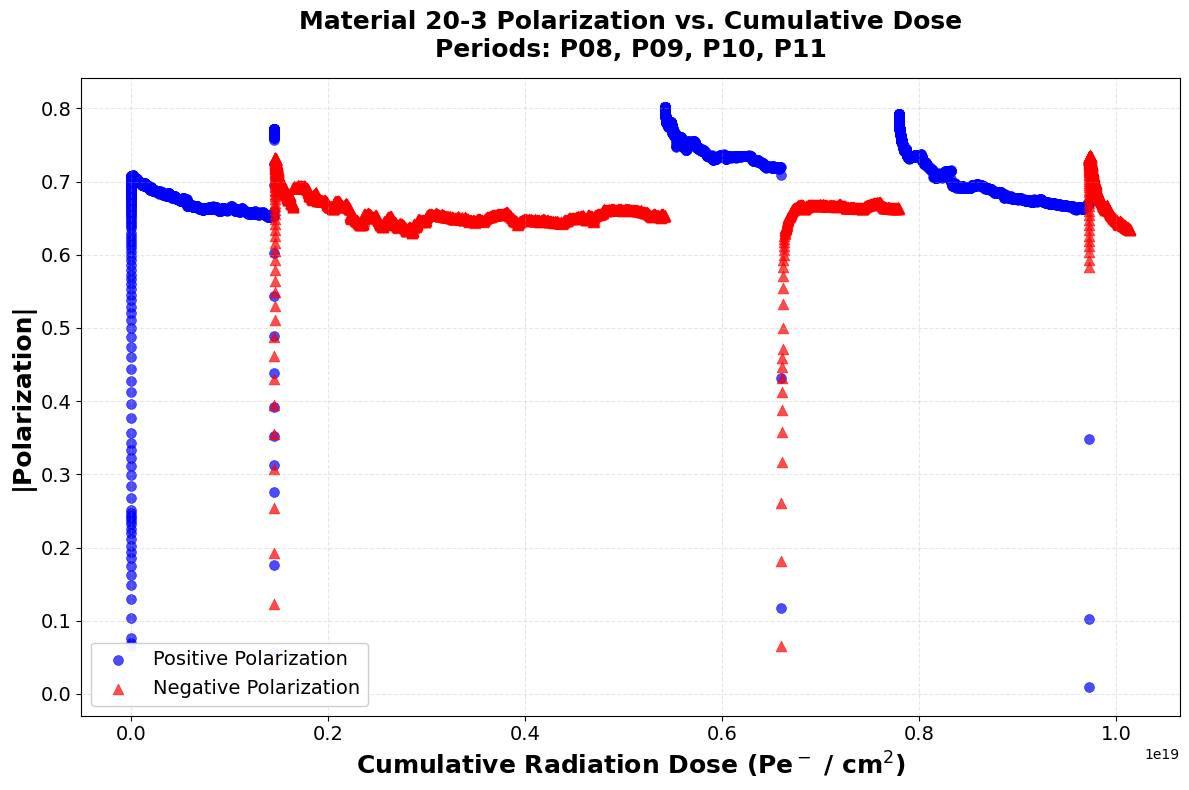}
    \caption{Proton polarization as a function of cumulative radiation dose for NH$_3$ sample \#20-3, used during data taking periods P08 through P11. The blue points correspond to positive polarization, and the red points correspond to negative polarization. In general, the rate at which the polarization degrades increases with the cumulative dose the material has received. The spikes in polarization can be due to either sudden changes in beam current or microwave frequency adjustments. }
    \label{fig:placeholder}
\end{figure}

Annealing the sample at temperatures around 100~K for several minutes can partially repair the radiation damage and recover much of the initial polarization. However, these results are diminishing, and the sample must be replaced after five to ten anneals.  This behavior is illustrated in Fig.\,\ref{fig:placeholder} which shows the increasingly rapid decay of polarization with cumulative dose.  The small spikes in polarization are due to microwave frequency adjustments to counter the decay.  



The exact microwave frequency needed at any given point in time depends on a number of hard-to-quantify factors, including the accumulated dose and annealing history of the sample.  In practice the frequency is determined by  adjusting the frequency until the polarization is maximized. Likewise, the rate at which the microwave frequency must change with the accumulation of dose is not precisely known, so maintaining an optimal polarization throughout an experiment depends on regular trial-and-error adjustments from a human operator. Maintaining optimal polarization is critical for the success of these experiments, as poor polarization directly impacts  the statistical significance of spin-dependent measurements. Currently, achieving and sustaining high polarization requires continuous monitoring and manual tuning by expert operators. 

For this study, we used a set of polarization data from the recent Run Group C experiments performed by the CLAS12 collaboration in Hall B, using the `APOLLO' (Ammonia POLarized LOngitudinally) target~\cite{Pandey:2024Wu}.




\subsection{Polarization Measurements}
A continuous-wave NMR system measures the target polarization for JLab's DNP solid targets \cite{maxwell_2026}. Using a device called a \textit{Q-Meter}, minute changes in the inductance of a coil surrounding the target are observed as an RF field is swept through the magnetic resonance frequency of the target nuclei. These changes reflect the bulk magnetization, and ultimately the polarization, of the sample.

\begin{figure}[htbp]
    \centering
    \includegraphics[width=1.0\linewidth]{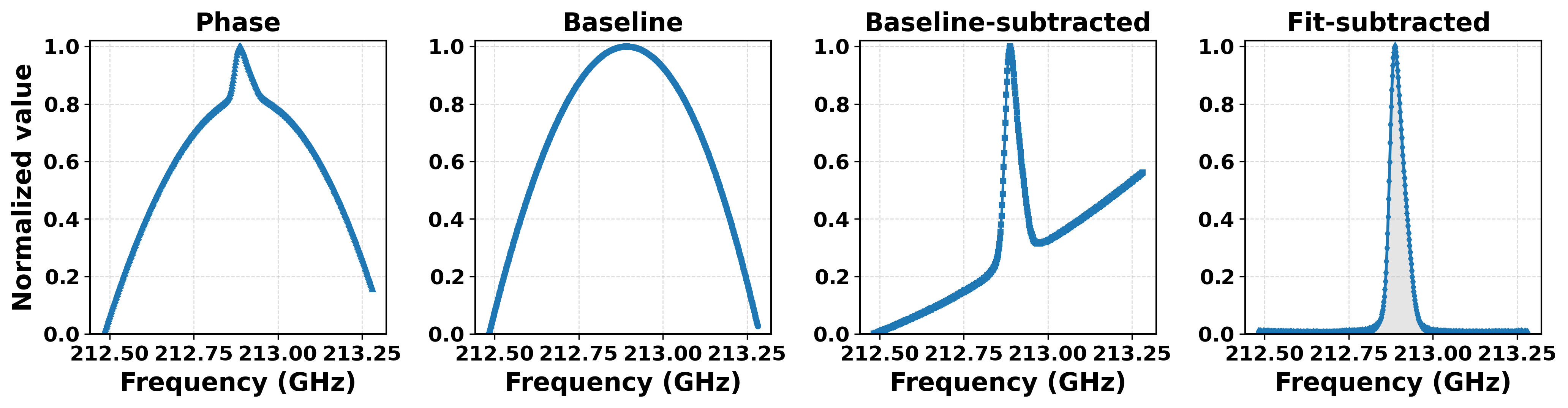}
    \caption{Example polarization extraction for protons in NH$_3$. The normalized phase, baseline, baseline-subtracted, and fit-subtracted signals versus frequency are shown. The polarization is proportional to the area under the fit-subtracted curve (shaded region). }
    \label{fig:p07_sample_pol}
\end{figure}

An example NMR measurement for an NH$_3$ target is shown in Fig. \ref{fig:p07_sample_pol}. The NMR absorption peak sits on top of a background ``Q-curve,'' which is the electronic circuit's voltage-versus-frequency response. While this curve is nearly parabolic, its true form is difficult to accurately model, so it is removed through subtraction. The background Q-curve is obtained by shifting the magnetic field so that the resonance frequency of the nuclei, and thus the NMR peak, is outside the frequency sweep range. The background signal is then subtracted from all subsequent measurements of the raw NMR (or ``phase'') signal.  The resulting curve may retain a slight degree of curvature resulting from temperature drifts of the NMR circuit.  This is removed using a polynomial fit to the signal wings.  The sample polarization is then extracted by integrating the area under the NMR peak and multiplying by a calibration constant. 


\subsection{Dataset \& Variable Selection }
A limitation of the current dataset is that individual microwave frequency adjustments by the shift crew are not logged as discrete events, only as changes in the measured microwave frequency. As a consequence, we cannot easily timestamp or attribute all operator interventions (outside of when the sign of the polarization is changed), which complicates comparisons between any automated system and humans.

\begin{figure}[htbp]
    \centering
    \includegraphics[width=1.0\linewidth]{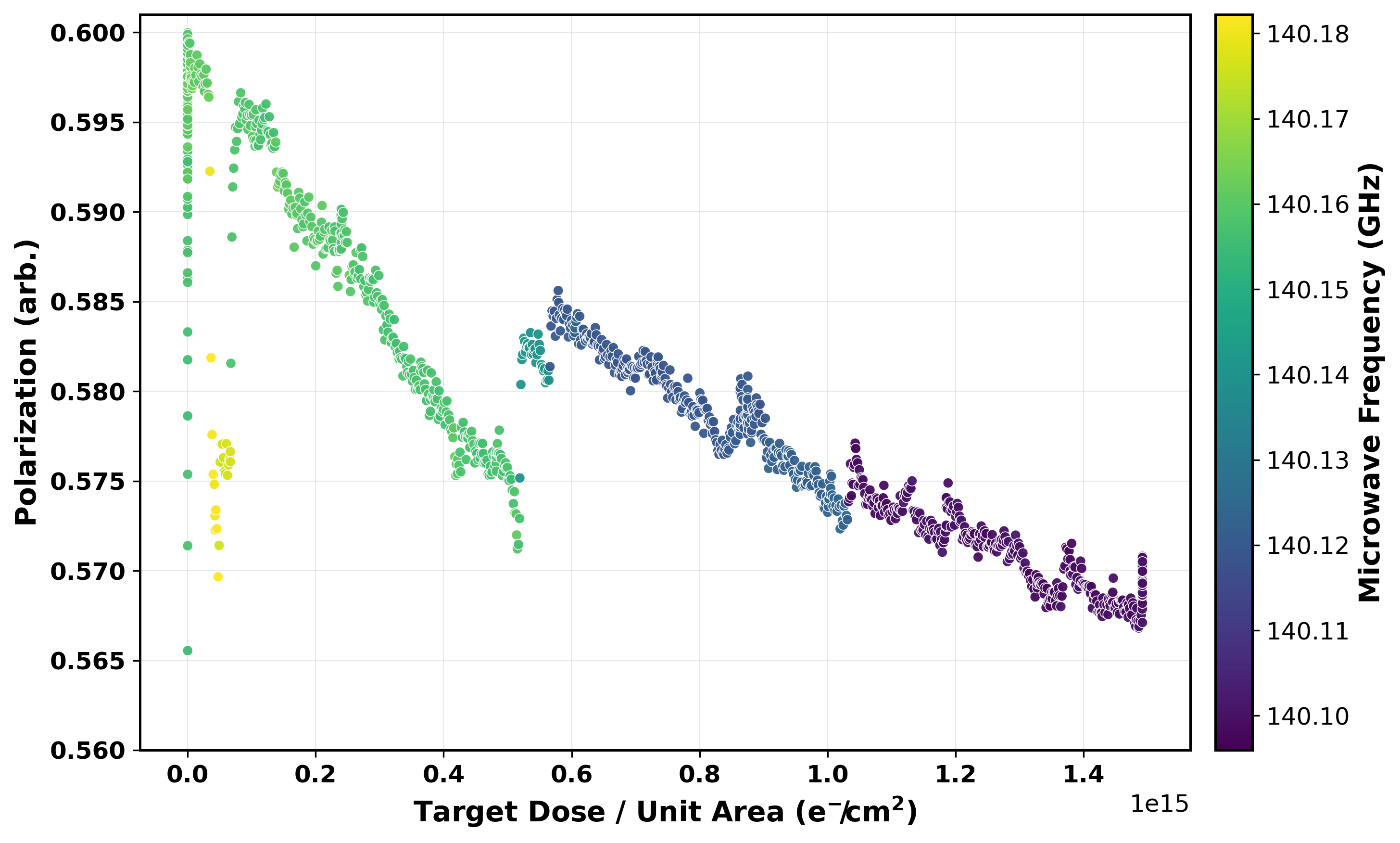}
    \caption{Sample response to changes in microwave frequency given the current polarization and dosage.}
    \label{fig:controls_environment_epics}
\end{figure}

\section{3 Previous Work}
\subsection{Automated Microwave Frequency Control}
The automatic adjustment of the microwave frequency could both improve the quality of data taken and reduce dependence on continued attention from fallible human operators. While there has been investigation into algorithmic and even machine-learning techniques for automated microwave frequency control, no such system has been successfully deployed in an experimental setting.

\section{4 Background}
\subsection{Data processing for ML}
This work uses a subset of data from run period P07 to develop single-sample models, comprising a total of 921 events. Each event contains multiple variables, including start time, end time, run number, microwave frequency, target dose, beam current, and polarization. Since the event duration (end time minus start time) is identical across all events, it was excluded from the feature set. Similarly, the run number, which serves only as an identifier for individual runs, was also excluded. Beam current, microwave frequency, and target dose per unit area were selected as input features, while polarization was treated as the output variable.

For the multi-sample models, data from run periods P07, P11, and P14 were used, resulting in a combined dataset of 5617 events. In both single-sample and multi-sample experiments, the dataset was divided into 80\% for model training and 20\% for testing.

To standardize the data, z-score normalization was applied to the input features. This transformation rescales each feature to have zero mean and unit variance and is defined as
\begin{equation}
z = \frac{x - \mu}{\sigma},
\label{eq:zscore}
\end{equation}
where $x$ denotes the original feature value, $\mu$ is the mean of the feature, and $\sigma$ is the corresponding standard deviation.

\subsection{Surrogate Models}
\subsection{MLPs}
\begin{figure}[htbp]
    \centering
    \includegraphics[width=1.0\linewidth]{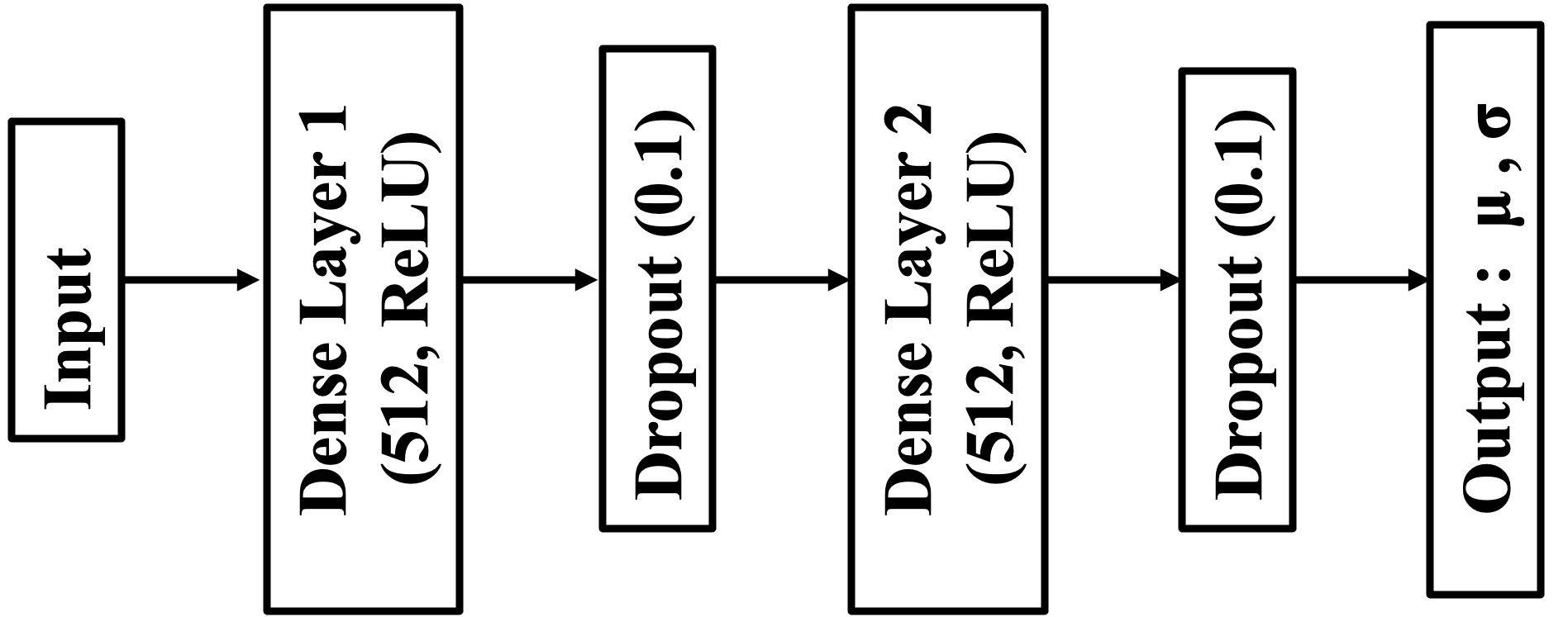}
    \caption{Schematic representation of the neural network architecture consisting of two dense hidden layers with ReLU activation and dropout regularization. The output layer predicts both the mean ($\mu$) and standard deviation ($\sigma$) of the target distribution for probabilistic regression.}

    \label{fig:block_MLP}
\end{figure}

In this study, we implemented a feedforward neural network comprising an input layer, two hidden layers, and an output layer. A schematic representation of the model architecture is presented in Figure~\ref{fig:block_MLP}. The input layer consists of three neurons, which correspond to the selected features. Each hidden layer contains 512 neurons. To mitigate overfitting, we incorporated dropout layers after each hidden layer. The output layer is composed of two neurons as the Negative Log-Likelihood (NLL) loss function requires the prediction of both the mean and variance of the target variable. The NLL loss function \cite{kendall2017uncertainties} is defined as follows:

\begin{equation} \label{eq:nll_no_pi}
\mathcal{L}_{\text{NLL}} = \frac{1}{2} \log \sigma^2 + \frac{(y - \mu)^2}{2\sigma^2}
\end{equation}

where $y$ represents the observed target value, $\mu$ denotes the predicted mean, and $\sigma^2$ corresponds to the predicted variance. Introducing the NLL loss function enables the model to learn both the expected value and the associated uncertainty of the target, thereby facilitating probabilistic regression.
\subsection{Gaussian Process (GP)}

A Gaussian Process (GP) is a non-parametric, probabilistic model used for regression and prediction tasks. GP models employ a kernel function to transform input data into a higher-dimensional representation \cite{schram2023uncertainty} that captures complex and non-linear relationships. The model makes predictions by utilizing the point-to-point distances between new inputs and training samples, where closer points have a greater influence on the prediction. Due to this property, GP models are considered distance-aware and can detect out-of-distribution (OOD) samples based on their distance from the training distribution\cite{schram2023uncertainty}.

This distance-aware behavior is governed by the choice of kernel function. Kernels, also referred to as covariance functions determine the structure of both the prior and posterior distributions of GP models. By defining a measure of similarity between input data points, kernels encode prior assumptions about the latent function being modeled. These assumptions govern the smoothness and correlation of the learned function and influence how the GP interpolates between observed data and quantifies uncertainty in regions with limited data.

Several kernel functions are commonly used in GP models, including the radial basis function (RBF), Laplace, and Cauchy kernels. These kernels are mathematically defined as \cite{rasmussen2006gaussian}

\begin{align}
k_{\text{RBF}}(\mathbf{x}_i,\mathbf{x}_j) &= \sigma_f^2 
\exp\!\left(-\frac{1}{2\ell^2}\|\mathbf{x}_i-\mathbf{x}_j\|_2^2\right), \\
k_{\text{Laplace}}(\mathbf{x}_i,\mathbf{x}_j) &= \sigma_f^2 
\exp\!\left(-\frac{1}{\ell}\|\mathbf{x}_i-\mathbf{x}_j\|_1\right), \\
\text{and}\quad
k_{\text{Cauchy}}(\mathbf{x}_i,\mathbf{x}_j) &= \sigma_f^2 
\left(1+\frac{\|\mathbf{x}_i-\mathbf{x}_j\|_2^2}{\ell^2}\right)^{-1}.
\end{align}

Here, $\sigma_f^2$ denotes the signal variance and $\ell$ is the characteristic length-scale. The vectors $\mathbf{x}_i,\mathbf{x}_j \in \mathbb{R}^d$ represent input feature vectors, while $\|\cdot\|_1$ and $\|\cdot\|_2$ denote the $\ell_1$ and $\ell_2$ norms, respectively. Each kernel function exhibits distinct characteristics in terms of smoothness, continuity, and correlation structure.

Let $\mathbf{X} \in \mathbb{R}^{N \times D}$ denote the matrix of training inputs and $\mathbf{X}^*$ the test inputs. The corresponding covariance matrices are defined as
\[
\mathbf{K} = K(\mathbf{X}, \mathbf{X}), \quad
\mathbf{K}_* = K(\mathbf{X}^*, \mathbf{X}), \quad
\mathbf{K}_{**} = K(\mathbf{X}^*, \mathbf{X}^*).
\]

The posterior predictive distribution at a test point $\mathbf{x}^*$ is Gaussian, with mean and variance given by \cite{rasmussen2006gaussian}

\begin{align}
\mu_* &= \mathbf{K}_*(\mathbf{K} + \sigma_n^2 \mathbf{I})^{-1}\mathbf{y}, \label{eq:gp_mean} \\
\sigma_*^2 &= \mathbf{K}_{**} - \mathbf{K}_*(\mathbf{K} + \sigma_n^2 \mathbf{I})^{-1}\mathbf{K}_*^\top. \label{eq:gp_variance}
\end{align}

The predictive mean in Eq.~\eqref{eq:gp_mean} is a linear combination of the observed training targets, while the predictive variance in Eq.~\eqref{eq:gp_variance} provides a quantitative measure of model uncertainty.

Deep neural networks (DNNs) are typically developed under the assumption that the training and test datasets come from the same distribution (i.e., In-Distributed, or ID). Such models often fail when the test data comes from OOD source. Ideally, a model should not only provide predictions but also indicate whether the input data is ID or OOD. GP models inherently provide this capability, as they can quantify uncertainty based on the distance from the training distribution. For example, Chen et al. \cite{chen2024uncertainty} presented a GP-based method that trains only on ID data and uses predictive uncertainty to distinguish between ID and OOD samples.

\subsection{Gaussian Process Approximation (GPA)}

Gaussian Process models are flexible probabilistic nonparametric methods that are widely used for regression, classification, and related tasks. A key advantage of GPs is that predictions are accompanied by principled uncertainty estimates. However, a major limitation of standard GP models is their computational intractability for large datasets. Specifically, GP training requires $\mathcal{O}(N^3)$ computational complexity due to inversion of the $N \times N$ covariance matrix, while prediction incurs an $\mathcal{O}(N^2)$ cost per test point, where $N$ denotes the number of training samples \cite{snelson2007local}.These computational demands grow rapidly with increasing $N$, making exact GP inference impractical for large-scale applications.

Gaussian Process Approximation (GPA) refers to a class of methods that approximate the GP kernel or posterior using a reduced representation. By replacing the full covariance structure with a lower-dimensional approximation, GPA significantly reduces computational cost while retaining the essential statistical properties of the original GP model.

Given input features $\mathbf{x} \in \mathbb{R}^d$, the kernel function $k(\mathbf{x}, \mathbf{x}')$ can be approximated using Random Fourier Features (RFF) as \cite{dai2020federated}

\begin{equation}
k(\mathbf{x}, \mathbf{x}') \approx \boldsymbol{\phi}(\mathbf{x})^\top \boldsymbol{\phi}(\mathbf{x}').
\end{equation}

The random feature map $\boldsymbol{\phi}(\mathbf{x}) \in \mathbb{R}^D$ is defined as \cite{dai2020federated}

\begin{equation}
\boldsymbol{\phi}(\mathbf{x}) = \sqrt{\frac{2}{D}} \cos\!\left(\mathbf{W}^\top \mathbf{x} + \mathbf{b}\right),
\end{equation}
where $D$ denotes the number of random features, $\mathbf{W} \in \mathbb{R}^{d \times D}$ is sampled from the spectral density of the kernel, and $\mathbf{b} \in \mathbb{R}^D$ is drawn element-wise from a uniform distribution $\mathcal{U}(0, 2\pi)$.

Using this approximation, Gaussian process inference can be performed in a finite-dimensional feature space. Let
\begin{equation}
\boldsymbol{\Phi}(\mathbf{X}_t) =
\begin{bmatrix}
\boldsymbol{\phi}(\mathbf{x}_1) \\
\boldsymbol{\phi}(\mathbf{x}_2) \\
\vdots \\
\boldsymbol{\phi}(\mathbf{x}_t)
\end{bmatrix}
\in \mathbb{R}^{t \times D}
\end{equation}
denote the random feature matrix constructed from the training inputs
$\mathbf{X}_t = \{\mathbf{x}_1, \ldots, \mathbf{x}_t\}$, and let
$\mathbf{y}_t \in \mathbb{R}^t$ be the corresponding observations. Assuming independent Gaussian observation noise with variance $\sigma^2$, the approximate GP posterior mean at a test input $\mathbf{x}$ is given by \cite{dai2020federated}

\begin{equation}
\hat{\mu}_t(\mathbf{x}) =
\boldsymbol{\phi}(\mathbf{x})^\top
\left(
\boldsymbol{\Phi}(\mathbf{X}_t)^\top \boldsymbol{\Phi}(\mathbf{X}_t)
+ \sigma^2 \mathbf{I}
\right)^{-1}
\boldsymbol{\Phi}(\mathbf{X}_t)^\top \mathbf{y}_t.
\end{equation}

The corresponding approximate posterior variance of the latent function is \cite{dai2020federated}

\begin{equation}
\hat{\sigma}_t^2(\mathbf{x}) =
\sigma^2 \,
\boldsymbol{\phi}(\mathbf{x})^\top
\left(
\boldsymbol{\Phi}(\mathbf{X}_t)^\top \boldsymbol{\Phi}(\mathbf{X}_t)
+ \sigma^2 \mathbf{I}
\right)^{-1}
\boldsymbol{\phi}(\mathbf{x}).
\end{equation}

\subsection{Reinforcement Learning}
\begin{figure}[h]
\centering
\includegraphics[width=0.46\textwidth]{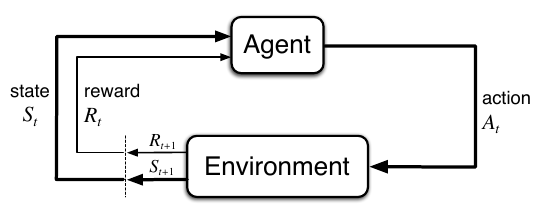}
\caption{The agent-environment interaction in a reinforcement learning workflow~\cite{sutton_barto_rl2018}.
The agent executes a policy that selects an action $A_t$ given the current $S_t$, which results in a reward $R_{t+1}$ and a new state $S_{t+1}$ of the environment.}
\label{fig:rl-model}
\end{figure}

Reinforcement Learning (RL) is a sub-field of AI focused on optimizing control or planning tasks by training an agent to interact with an environment and learn from feedback. As illustrated in Figure~\ref{fig:rl-model}, at each timestep $t$, the agent observes the environment state $S_t$, takes an action $A_t$, and receives a reward $R_{t+1}$ while the environment transitions to a new state $S_{t+1}$. The agent's goal is to maximize cumulative long-term reward, which can be expressed using the Q-value function $Q(S,A)$ via the Bellman equation ~\ref{eq:bellman}:

\begin{equation}
\label{eq:bellman}
    Q(S, A) = E\left[ R_{t+1} + \gamma \max_{A'} Q(S', A')\right]
\end{equation}

Here, $\gamma$ is a discount factor for future rewards, and $Q(S',A')$ represents the maximum expected reward from the next state $S'$. In Deep Reinforcement Learning (DRL), the agent and the Q-function (critic) are modeled using deep neural networks and are trained simultaneously.

A variety of DRL algorithms exist for control tasks. For this work, we use the Twin-Delayed Deep Deterministic Policy Gradient (TD3) algorithm ~\cite{rl_td3}. TD3 is an actor-critic method that employs two critic networks to reduce overestimation of Q-value, stabilizing training. The agent interacts with the environment, storing experiences $(S_i, A_i, S_{i+1}, R_{i+1})$ in a replay buffer. Mini-batches sampled from this buffer are used to update the actor and critic networks iteratively. The reward function for TD3 is designed using trajectory references, evaluating how an action $A$  at state $S$ aligns the system with a target trajectory. Additional terms, such as penalties for large actions, can be incorporated to encourage smooth control. To safely train the RL agent, the surrogate model of the physical system is wrapped as a standard RL environment using the gymnasium package. This allows the agent to explore and optimize control policies in a virtual setting before deployment on the actual system. Once trained, the policy can be deployed to maintain or optimize target states in real experiments. 

\noindent

\section{5 Methods}

\subsection{Surrogate Model of Target }

\subsection{Multilayer Perceptron (MLP)}

\begin{figure}[htbp]
    \centering
    \includegraphics[width=1.0\linewidth]{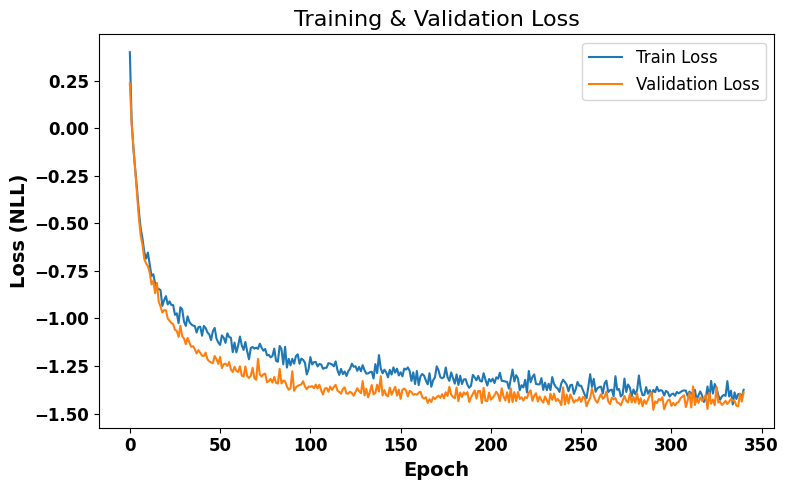}
    \caption{ Learning curves showing the training and validation loss for the MLP model. The validation loss decreases as the number of epochs increases, and training stops when the minimum validation loss is reached.}
    \label{fig:Learning}
\end{figure}
In this work, MLP model was developed using the TensorFlow framework and trained with the Adam optimizer at a learning rate of 0.0001. We applied early stopping with a patience of 25 epochs to prevent overfitting. The learning curve, shown in Figure ~\ref{fig:Learning}, demonstrates that both the training and validation losses decreased steadily with increasing epochs, indicating effective learning.
\begin{figure}[htbp]
    \centering
    \includegraphics[width=1.0\linewidth]{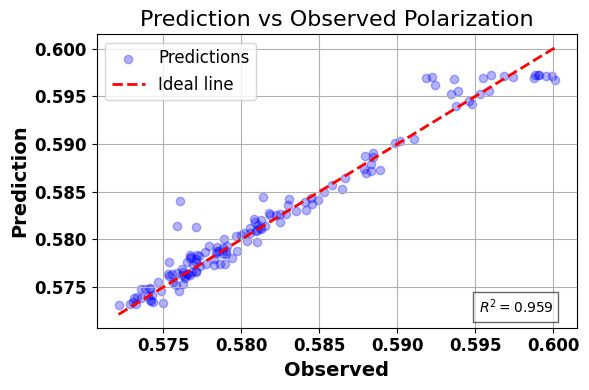}
    \caption{ Observed and predicted polarization for the MLP model. The red dotted line indicates the ideal prediction, and the blue markers represent the model’s predictions.}
    \label{fig:Pred_GT_MLP}
\end{figure}

\hspace{1em} We evaluated the performance of our trained model on the test dataset reserved during the data split. Figure~\ref{fig:Pred_GT_MLP} shows the comparison between the observed and predicted polarization values. The blue markers represent the predicted polarization, while the red dotted line indicates the ideal one-to-one relationship. Most of the predicted values lie close to the ideal line, with only a few deviations. To quantify the performance, we calculated the coefficient of determination ($R^2$) between the observed and predicted values. An $R^2$ value close to 1 indicates an excellent fit, while a value near 0 suggests a poor fit. Our calculated $R^2$ value of 0.959 demonstrates that the model explains approximately 95.9\% of the variance in the data.
\begin{figure}[htbp]
    \centering
    \includegraphics[width=1.0\linewidth]{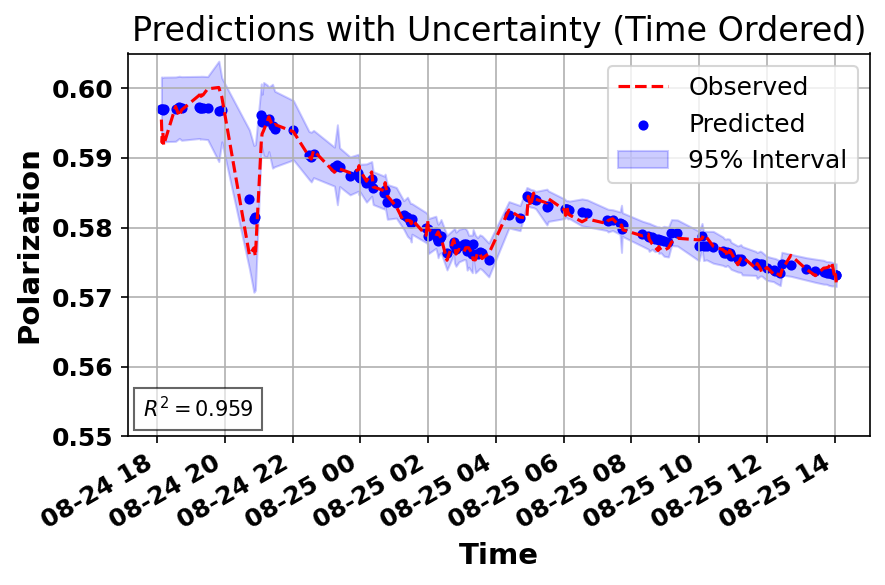}
    \caption{ Sample-wise observed values, predictions, and prediction intervals. The red dotted line represents the observed values, and the blue markers represent the model predictions. The predictions follow patterns similar to the true values, with associated uncertainty.}
    \label{fig:MLP_Prediction}
\end{figure}

Figure ~\ref{fig:MLP_Prediction} shows the observed polarization, the mean predicted polarization, and the corresponding prediction intervals for each test sample. The plot indicates that the mean predictions closely follow observed polarization values. It is also observed that the observed polarization generally lies within the prediction intervals. 

\begin{figure}[htbp]
    \centering
    \includegraphics[width=1.0\linewidth]{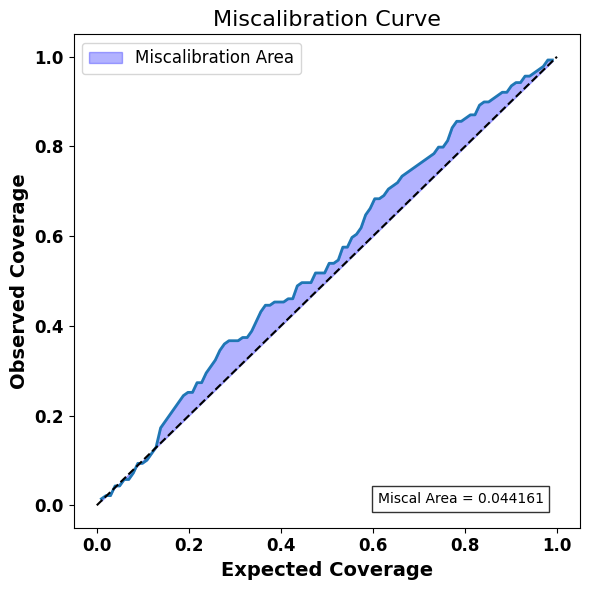}
    \caption{ Average calibration of predicted vs. observed values for the MLP model.}
    \label{fig:Miscalibration_MLP}
\end{figure}

The uncertainty toolbox \cite{chung2021uncertainty} was used to calibrate the uncertainty. Figure~\ref{fig:Miscalibration_MLP} presents the average calibration across the test data. A perfectly calibrated model would produce a curve indicated by the dotted yellow line. The deviation between the model’s calibration curve and the ideal line can be quantified using the miscalibration area, with a larger area representing a greater discrepancy between observed and predicted uncertainty. A curve below the ideal line indicates overconfidence, while a curve above the ideal line indicates underconfidence. In this case, the model’s calibration curve lies above the ideal line, indicating underconfidence. The MLP model’s miscalibration area is 0.044.

\subsection{Gaussian Process (GP)}

\begin{figure}[htbp]
    \centering
    \includegraphics[width=1.0\linewidth]{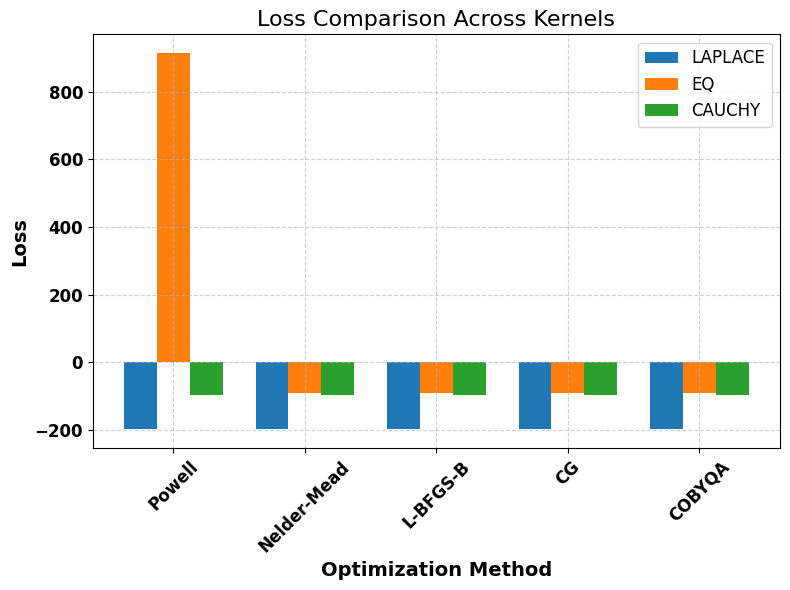}
    \caption{ Optimized NLL loss values for Laplace, RBF, and Cauchy kernels using Powell, Nelder–Mead, L-BFGS-B, CG, and COBYQA optimization methods.}
    \label{fig:NLL_loss_GP}
\end{figure}

In this work, we applied three different kernels, Radial Basis Function (RBF), Laplace, and Cauchy, to perform polarization prediction using GP models. The performance of these GP models is primarily determined by three kernel parameters: length scale, coefficient, and noise. To obtain the optimal set of parameters for our dataset, we minimized the NLL loss function. The optimization was performed using the SciPy minimizer with five different optimization algorithms: Powell, Nelder–Mead, Limited-memory Broyden–Fletcher–Goldfarb–Shanno Bound-constrained (L-BFGS-B), Conjugate Gradient (CG), and Constrained Optimization BY Quadratic Approximations (COBYQA). Across all kernels, the results obtained from different optimizers were largely consistent. Figure ~\ref{fig:NLL_loss_GP} illustrates the NLL loss for each optimizer across the three kernels, showing that the Laplace kernel achieved the lowest NLL loss compared to the RBF and Cauchy kernels. 

\begin{figure}[htbp]
    \centering
    \includegraphics[width=1.0\linewidth]{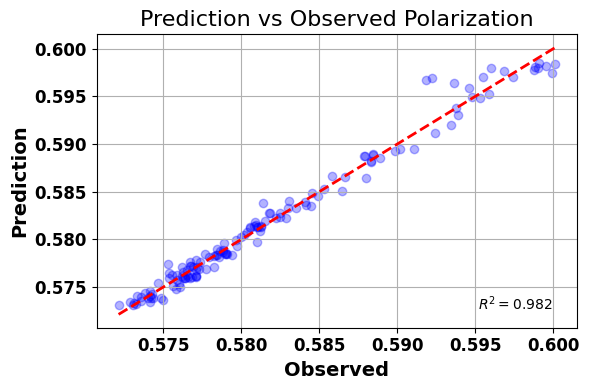}
    \caption{ Observed and predicted polarization for the GP model. The red dotted line indicates the ideal prediction, and the blue markers represent the model’s predictions.}
    \label{fig:Pre_obs_GP}
\end{figure} 

Figure ~\ref{fig:Pre_obs_GP} presents the scatter plot of observed vs. predicted polarization for the GP model using the Laplace kernel and corresponding optimized hyperparameters. Similar to the MLP plot in Figure~\ref{fig:Pred_GT_MLP} , the red dotted line represents the ideal performance. The GP model performs very closely to the ideal line in most cases, with only a few samples showing small deviations from the expected values. The R2 value between the observed and predicted polarization is 0.982, indicating that the model captures 98.2\% of the variation of the observed test data.

\begin{figure}[htbp]
    \centering
    \includegraphics[width=1.0\linewidth]{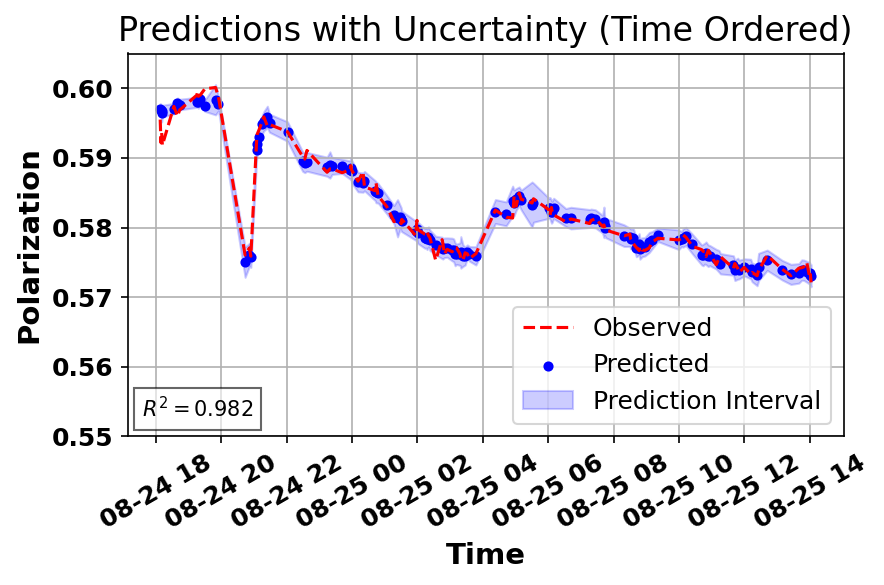}
    \caption{ Sample-wise observed values, predictions, and prediction intervals. The red dotted line represents the observed values, and the blue markers represent the model predictions. The predictions follow patterns similar to the true values, with associated uncertainty.}
    \label{fig:prediction_interval_GP}
\end{figure} 

Observed polarization, predicted polarization, and prediction intervals (about 95\% confidence interval) are presented in Figure ~\ref{fig:prediction_interval_GP} for the GP model on the test data. The horizontal axis represents time, and the vertical axis denotes polarization. The observed polarization changes over time, while the GP model predictions follow similar patterns to the observed values. The prediction interval illustrates the uncertainty of the predictions, with the observed values falling within this range for most cases.

\begin{figure}[htbp]
    \centering
    \includegraphics[width=1.0\linewidth]{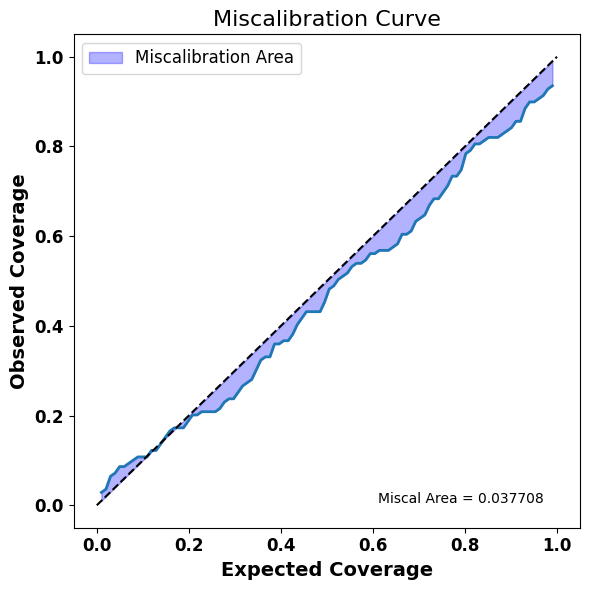}
    \caption{Average calibration of predicted vs. observed values for the MLP model.}
    \label{fig:Miscalibration_GP}
\end{figure} 
Average calibration across the test data is presented in  Figure~\ref{fig:Miscalibration_GP} for the GP model. The curve crosses the ideal diagonal, indicating that the model is overconfident in some ranges of predictions and underconfident in others. The miscalibration area is 0.037, reflecting a high degree of agreement between observed and predicted uncertainty.

\subsection{Gaussian Process Approximation (GPA)}
%
We developed GP and MLP models using data from a single run period (P07), consisting of fewer than 1000 data points. The models use sample-specific input variables, including microwave frequency, beam current, and target dose, to predict the polarization output. Since both the input and output variables span limited ranges, the resulting models exhibit limited generalization capability to other sample IDs. This limitation also poses challenges for reinforcement learning agents, as it restricts their ability to effectively select the control variable (microwave frequency) to achieve optimized polarization. To improve generalization and enable more robust control, it is therefore necessary to develop a multi-sample model trained on data from multiple sample IDs.

In this work, data from multiple run periods (P07, P11, and P14) are used to construct the Gaussian Process Approximation (GPA) model. As a first step, the Random Fourier Feature (RFF) approximation is employed to sample random frequencies from a distribution determined by the selected kernel. These frequencies are used to project the input training data into a finite-dimensional feature space, where cosine-based random features with random phase shifts and appropriate normalization are constructed. This transformation maps the original inputs into a feature representation whose inner products approximate the kernel function.

Based on the resulting random feature representation, the Gaussian process hyperparameters, including the length scale, kernel coefficient, and noise variance, are learned by maximizing the marginal likelihood of the approximate GP model. Using the learned hyperparameters, the final random feature mapping is constructed, and the corresponding regression weights are obtained by solving a regularized linear system. This enables efficient prediction within the reduced feature space. The predictive mean is computed as a linear combination of the random features, while predictive uncertainty is quantified by propagating the observation noise through the regularized feature-space covariance.

\begin{figure}[htbp]
    \centering
    \includegraphics[width=1.0\linewidth]{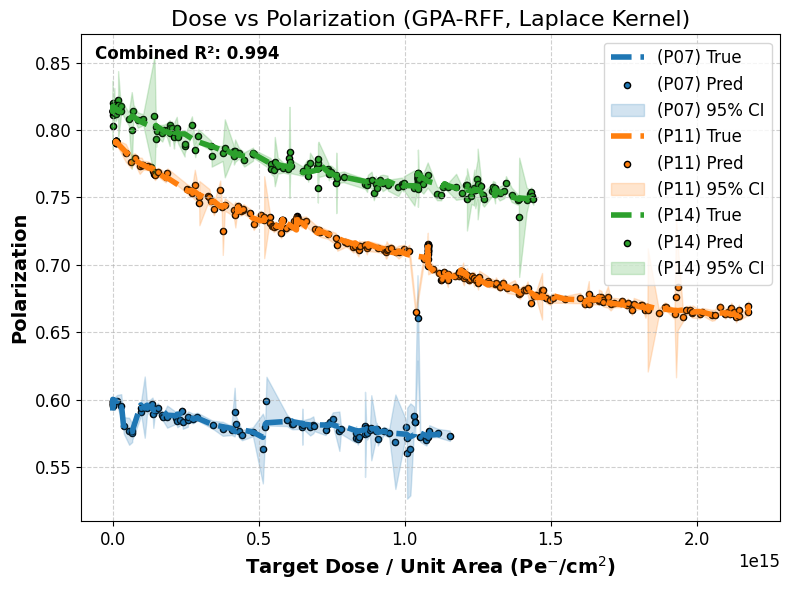}
    \caption{Sample-wise comparison of observed polarization values and GPA model predictions for the P07, P11, and P14 test datasets. Solid lines indicate the observed values, while markers denote the model predictions. The shaded regions represent the 95\% predictive confidence intervals for each individual sample ID.}
    \label{fig:prediction_interval_GPA}
\end{figure}

\begin{figure}[htbp]
    \centering
    \includegraphics[width=1.0\linewidth]{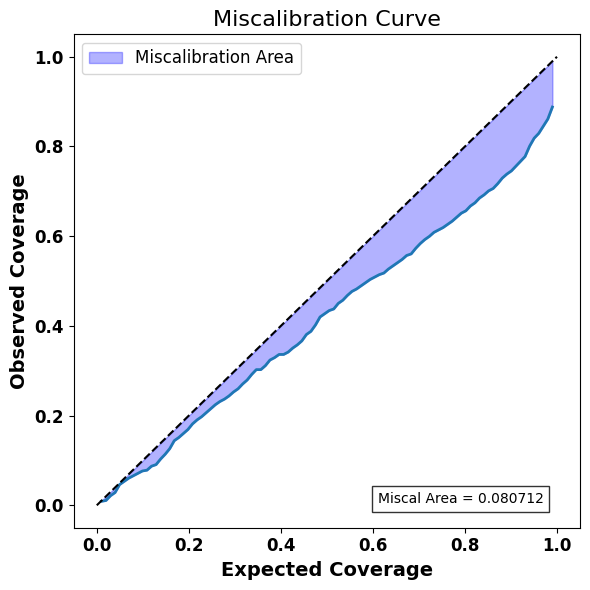}
    \caption{Average calibration of predicted vs. observed values for the GPA model for P07, P11, and P14.}
    \label{fig:Miscalibration_GPA}
\end{figure}

Figure~\ref{fig:prediction_interval_GPA} illustrates the polarization predictions obtained using a Gaussian kernel--based GPA model trained on multiple sample IDs. The model achieves an $R^2$ value of 99.4\% on the test dataset across all three samples. Figure~\ref{fig:Miscalibration_GPA} presents the corresponding miscalibration area plot, where the generalized model exhibits a miscalibration area of 0.08 on the same test data, indicating its uncertainty calibration performance.

\subsection{Out-of-Distribution (OOD) Behavior Analysis}
Figure~\ref{fig:MLP_Prediction} and Figure~\ref{fig:prediction_interval_GP} present the performance of the MLP and GP models on the test dataset, respectively. Both figures illustrate the model predictions along with their corresponding uncertainties for ID data. When comparing the two models, the MLP shows higher uncertainty compared to the GP. 

To analyze the OOD behavior, we examined the model responses when the microwave frequency was varied beyond the range observed in the training data. In our dataset, the training frequency spans from 140.0979~GHz to 140.1822~GHz. For this analysis, the frequency was kept identical to the test data up to a target dose of $0.4 \times 10^{15}$ (an arbitrary value selected to observe the transition from ID to OOD), after which it was gradually increased and decreased to move beyond the training range. All other input parameters were taken from their actual measured values in the P07 dataset and were kept unchanged; only the microwave frequency was modified for this analysis.

\begin{figure*}[htbp]
    \centering

    \begin{subfigure}[b]{0.49\textwidth}
        \centering
        \includegraphics[width=\textwidth]{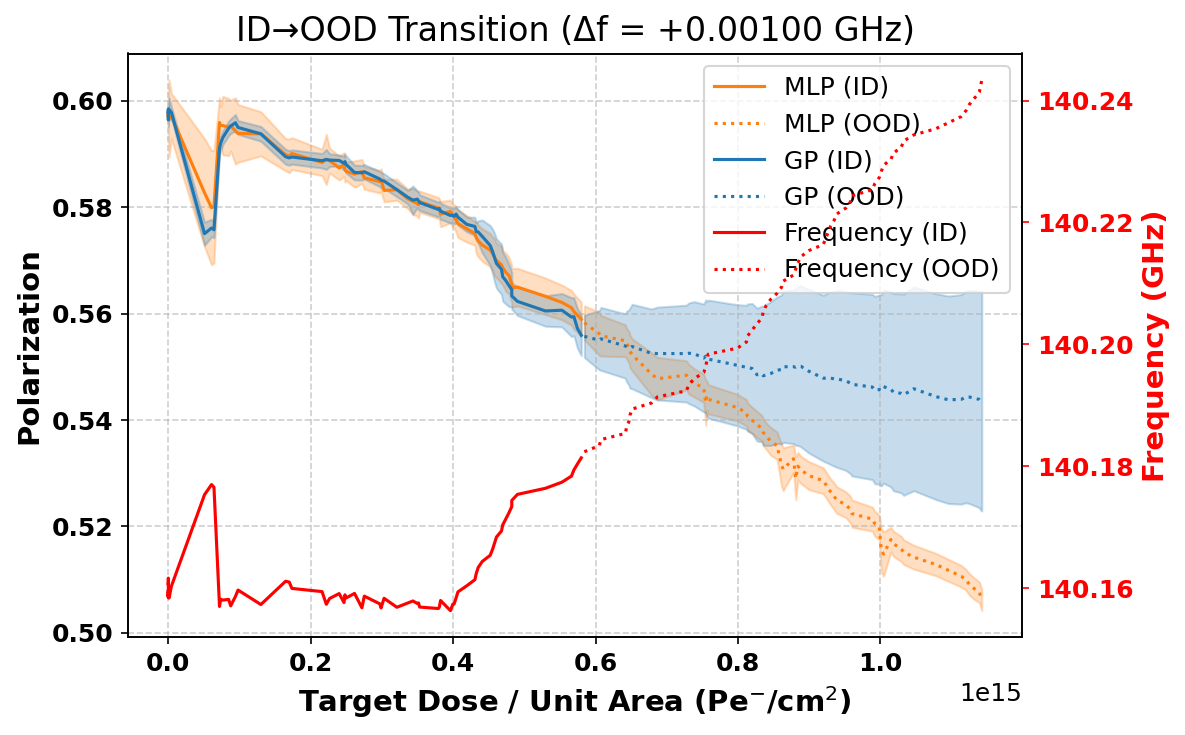}
        \caption{}
        \label{fig:OOD_above}
    \end{subfigure}
    \hfill
    \begin{subfigure}[b]{0.49\textwidth}
        \centering
        \includegraphics[width=\textwidth]{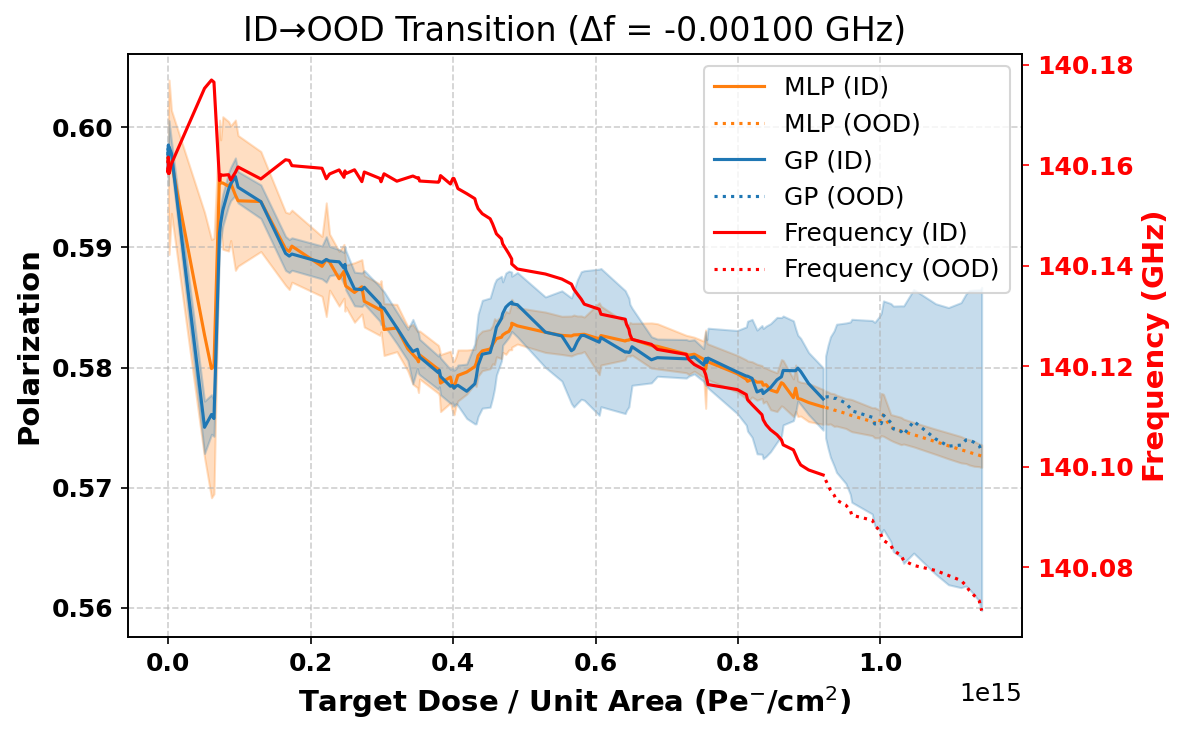}
        \caption{}
        \label{fig:OOD_below}
    \end{subfigure}

    \vspace{0.5em}

    \begin{subfigure}[b]{0.49\textwidth}
        \centering
        \includegraphics[width=\textwidth]{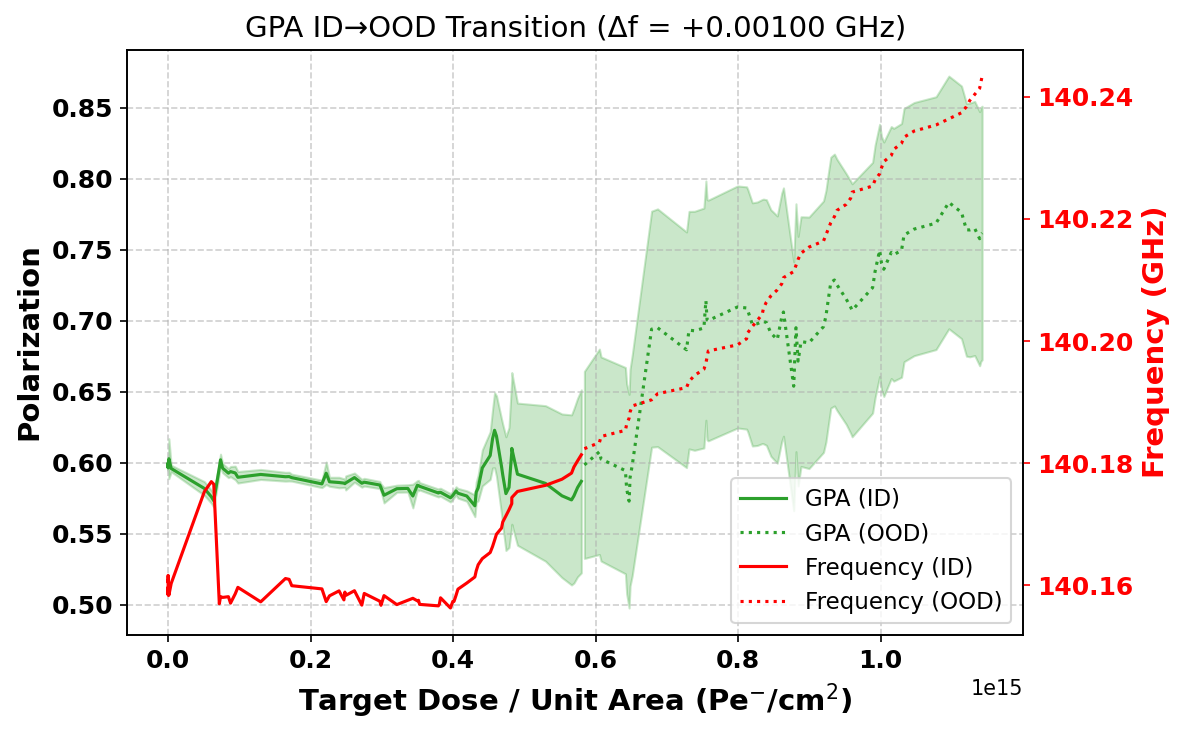}
        \caption{}
        \label{fig:OOD_GPA_above}
    \end{subfigure}
    \hfill
    \begin{subfigure}[b]{0.49\textwidth}
        \centering
        \includegraphics[width=\textwidth]{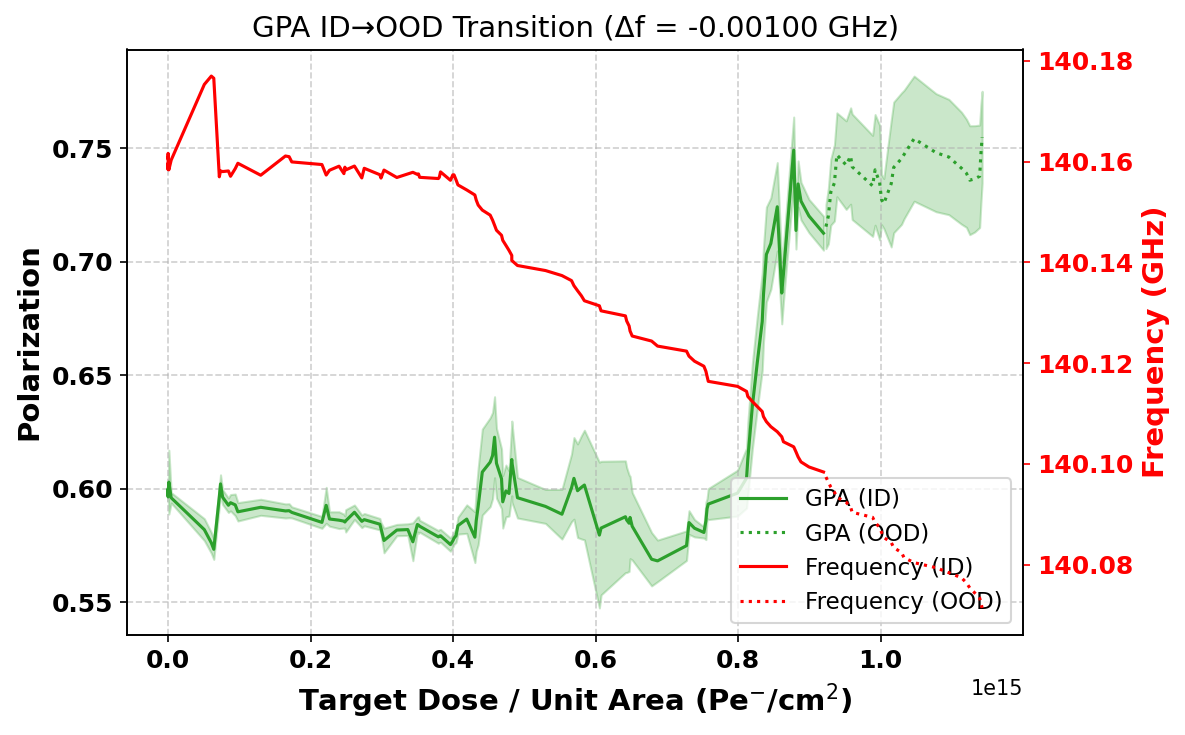}
        \caption{}
        \label{fig:OOD_GPA_below}
    \end{subfigure}

\caption{
Comparison of MLP, GP, and GPA responses during controlled ID-to-OOD transitions. The MLP and GP models are trained using single-sample P07 data, while the GPA model is trained using multisample data from samples P07, P11, and P14. All models are evaluated on P07 test data, with microwave frequency intentionally varied from ID (solid red line) to OOD region (dashed red line). Shaded regions represent prediction uncertainty. In (a) and (c), the frequency is increased beyond the upper training limit of $140.1822$ GHz, while in (b) and (d), it is decreased below the lower limit of $140.0979$ GHz. 
}
    \label{fig:OOD_combined_all}
\end{figure*}

Figures~\ref{fig:OOD_combined_all}(a) and~\ref{fig:OOD_combined_all}(b) show the GP and MLP responses for the ID and OOD regions. Solid lines represent the ID response, while dotted lines represent the OOD response, with shaded regions indicating prediction uncertainty. The MLP produces predictions with moderate uncertainty but does not clearly distinguish between the ID and OOD regions. In contrast, the GP exhibits substantially higher uncertainty in the OOD regime. As the frequency moves farther beyond the training range, the GP uncertainty continues to increase, providing a clear indication of the OOD region.

Figures~\ref{fig:OOD_combined_all}(c) and~\ref{fig:OOD_combined_all}(d) show the responses of the GPA model, which was trained using data from three sample IDs, P07, P11, and P14, and evaluated on P07 test data. Solid and dotted lines represent the ID and OOD responses, respectively, with shaded regions indicating prediction uncertainty. Before the intentionally introduced frequency variation, the GPA model produces well-behaved predictions with relatively small uncertainty. All other input variables retain their actual P07 test values, so the microwave frequency is the only variable intentionally varied beyond the training distribution. The ID behavior is consistent with that observed for the GP model. As the frequency moves beyond the training range into the OOD regime, the GPA uncertainty increases substantially, clearly indicating the distribution shift. This result demonstrates that the GPA retains sensitivity to frequency-based OOD behavior even when trained using data from multiple sample IDs.

\subsection{Simulated Environments}

Effective algorithm training and testing requires a standardized environment that interfaces with control software packages and enables smooth transitions between simulation and real hardware. We use Gymnasium \cite{towers2025gymnasiumstandardinterfacereinforcement}, the industry-standard open-source framework maintained by The Farama Foundation, which provides the infrastructure to encapsulate simulations with consistent state spaces, action spaces, and reward structures to preserve policy transferability.

We employ two environment versions based on data-driven surrogate models with uncertainty quantification: a single sample and a multi sample variant. Both maintain identical action and observation spaces but differ in how they estimate model uncertainty.

\subsubsection{Action Space}

Each environment version supports both discrete and continuous action spaces. The discrete space as seen in Table \ref{tab:discaction} mirrors the three options currently available to counting house operators: do nothing, increase, or decrease the microwave frequency. The continuous space as seen in Table \ref{tab:contaction} provides more granular frequency control within the same range. For the purposes of this paper, we use the continuous action space to give the reinforcement learning agent the best opportunity at finding an optimal solution.

\begin{table}[h!]
\centering
\caption{Discrete Action Space for P07 Environment}
\label{tab:discaction}
\begin{tabular}{|c|l|c|}
\hline
\textbf{Num} & \textbf{Name} & \textbf{$\Delta$UWave} \\
\hline
0 & No action & 0 \\
1 & Decrease UWave Frequency & $-0.001$ \\
2 & Increase UWave Frequency & $+0.001$ \\
\hline
\end{tabular}
\end{table}

\begin{table}[h!]
\centering
\caption{Continuous Action Space for P07 Environment}
\label{tab:contaction}
\begin{tabular}{|l|l|c|}
\hline
\textbf{Num} & \textbf{Name} & \textbf{$\Delta$UWave} \\
\hline
0 & UWave Frequency Adjustment & $[-0.01, +0.01]$ \\
\hline
\end{tabular}
\end{table}

\subsubsection{Observation Space}
The observation space is an 8-dimensional vector designed to enable uncertainty-aware decision making and smooth control, summarized in Table~\ref{tab:obs_space}.

\begin{table}[htb!]
\centering
\caption{Observation Space for Proton Polarization Environment}
\label{tab:obs_space}
\begin{tabular}{|c|l|c|c|}
\hline
\textbf{Num} & \textbf{Name} & \textbf{Min} & \textbf{Max} \\
\hline
0 & Normalized Frequency & $0.0$ & $1.0$ \\
1 & Scaled Target Dose & $0.0$ & $\approx 4.0$ \\
2 & Beam Current & $0.0$ & $\approx 5.0$ \\
3 & Predicted Mean Polarization & $0.0$ & $1.0$ \\
4 & Predicted Std. Deviation & $0.0$ & $0.2$ \\
5 & Lower Confidence Bound & $0.0$ & $1.0$ \\
6 & Frequency Derivative & $-5.0$ & $5.0$ \\
7 & Previous Action & $-1.0$ & $1.0$ \\
\hline
\end{tabular}
\end{table}

The first three dimensions capture the physical system state. The microwave frequency is normalized to $[0, 1]$ based on the allowable bounds $[f_{\text{min}}, f_{\text{max}}]$ (typically [139.5, 140.25] GHz) to aid neural network learning. The target dose and beam current, both determined by historical P07 run data, are scaled (by $10^{15}$ and raw values, respectively) to maintain reasonable numerical ranges.

Dimensions 3--5 provide the surrogate model's polarization predictions and a derived uncertainty metric. The predicted mean $\mu_t$ and standard deviation $\sigma_t$ are included directly, along with the Lower Confidence Bound $\text{LCB}_t = \mu_t - k\sigma_t$ (typically $k=3$). Including the LCB explicitly allows the agent to observe the trade-off between high polarization and low uncertainty that drives the reward function.

The final two dimensions encode temporal information: the normalized rate of frequency change $(f_t - f_{t-1})/\Delta f_{\text{scale}}$ and the previous action $a_{t-1}$. These enable the agent to learn smooth control policies and avoid oscillatory behavior.

\subsubsection{Reward Structure}
The reward function uses a Lower Confidence Bound (LCB) formulation that balances high polarization predictions against model uncertainty, while penalizing excessive control actions.

\paragraph{Core Reward.}
The primary reward component is the LCB of the surrogate model's polarization prediction:
\begin{equation}
\text{LCB}_t = \mu_t - k \sigma_t
\end{equation}
where $\mu_t$ is the predicted mean polarization, $\sigma_t$ is the epistemic uncertainty, and $k$ is the confidence multiplier (typically $k=3$, corresponding to $\approx$99.7\% confidence under Gaussian assumptions). This naturally encourages operating points where predicted polarization is high and uncertainty is low. For stable learning, the LCB is normalized to approximately $[0, 1]$:
\begin{equation}
\text{LCB}_{\text{norm}} = \text{clip}\left(\frac{\text{LCB}_t - 0.4}{0.2}, 0, 1.5\right)
\label{eq:LCBNORM}
\end{equation}
where the normalization assumes LCB values typically range from 0.4 to 0.6 based on the observed data distribution, and clipping above 1.0 allows bonus rewards.

\paragraph{Regularization Penalties.}
Two penalty terms promote smooth, efficient control. An action penalty $r_{\text{action}} = -\lambda_a \cdot a_t^2$ discourages extreme actions, where $a_t \in [-1, 1]$ is the normalized action and $\lambda_a = 0.05$. A frequency change penalty $r_{\text{freq}} = -\lambda_f \cdot |({f_t - f_{t-1}})/{\Delta f_{\text{scale}}}|$ encourages smooth frequency trajectories, where $\lambda_f = 0.1$.

\paragraph{Total Reward.}
The complete step reward and episodic return are:
\begin{equation}
r_t = \alpha \cdot \text{LCB}_{\text{norm}} - \lambda_a \cdot a_t^2 - \lambda_f \cdot \left|\frac{f_t - f_{t-1}}{\Delta f_{\text{scale}}}\right|
\end{equation}
\begin{equation}
R_{\text{episode}} = \sum_{t=0}^{T-1} r_t
\end{equation}
where $\alpha = 1.0$ is the reward scale parameter and $T$ is the number of time steps in the episode.

\section{Results}

\begin{figure}[htbp]
    \centering
    \includegraphics[width=1.0\linewidth]{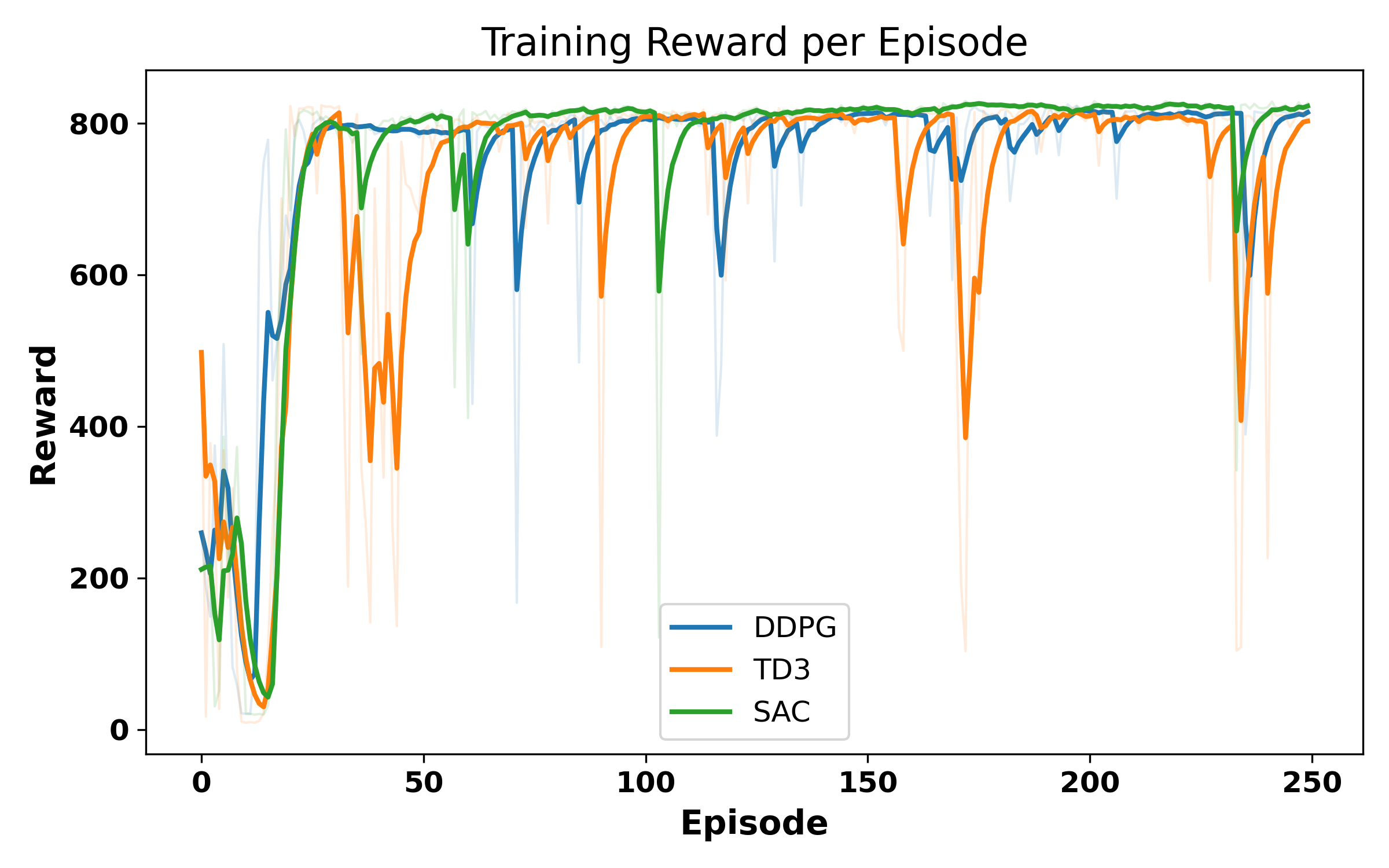}
    \caption{ Episode reward curves for DDPG, TD3, and SAC. All three algorithms converge to comparable reward values, indicating that final performance is largely independent of the specific off policy actor-critic algorithm used.}
    \label{fig:algorithmcomparison}
\end{figure}

Three actor-critic algorithms were evaluated: Deep Deterministic Policy Gradient (DDPG) \cite{Silver2014DeterministicPG}, Twin Delayed Deep Deterministic Policy Gradient (TD3) \cite{td3}, and Soft Actor-Critic (SAC) \cite{sac}. All three converged to similar reward values across the models tested as seen in Fig. \ref{fig:algorithmcomparison}, indicating that the choice of algorithm was less significant than the environment and reward formulation. TD3 was selected for the studies presented below. A differential version of the TD3 algorithm (Diff-TD3, outlined in Alg. \ref{alg:diff_td3}) was also evaluated in an attempt to reduce the training time and sample efficiency while utilizing the multi sample GPA models in the environment.

Default hyperparameters for the TD3 agent are listed in Table~\ref{tab:td3-hyperparameters}. Tanh was used as the final activation function to bound the actor's output to $[-1, 1]$, enabling smooth transformation to the environment's action space. The following results section is split into two with the single sample GP model results being presented first which lead into the multi sample GPA results.

\begin{table}[h]
\centering
\caption{Default hyperparameters for the TD3 algorithm.}
\label{tab:td3-hyperparameters}
$$
\begin{array}{|l|c|}
\hline
\textbf{Hyperparameter} & \textbf{Value} \\
\hline
\text{Warmup Size} & 2500 \\
\text{Batch Size} & 256 \\
\text{Critic Learning Rate} & 0.0003 \\
\text{Actor Learning Rate} & 0.0003 \\
\text{Tau (Soft Update)} & 0.005 \\
\text{Discount Factor } (\gamma) & 0.99 \\
\text{Buffer Capacity} & 1{,}000{,}000 \\
\text{Buffer Type} & \text{ER} \\
\text{Actor Model} & \text{FCNN} \\
\text{Critic Model} & \text{FCNN} \\
\text{Exploration Noise Fraction} & 0.1 \\
\hline
\textbf{Network Architecture} & \textbf{A\&C} \\
\hline
\text{Hidden Layers} & 2 \\
\text{Nodes per Layer} & [256, 256] \\
\text{Activation Functions} & [\text{ReLU}, \text{ReLU}] \\
\text{Final Activation} & [\text{Tanh}] \\
\hline
\end{array}
$$
\end{table}

\subsection{Single Sample Model}

\begin{figure}[htbp]
    \centering
    \includegraphics[width=1.0\linewidth]{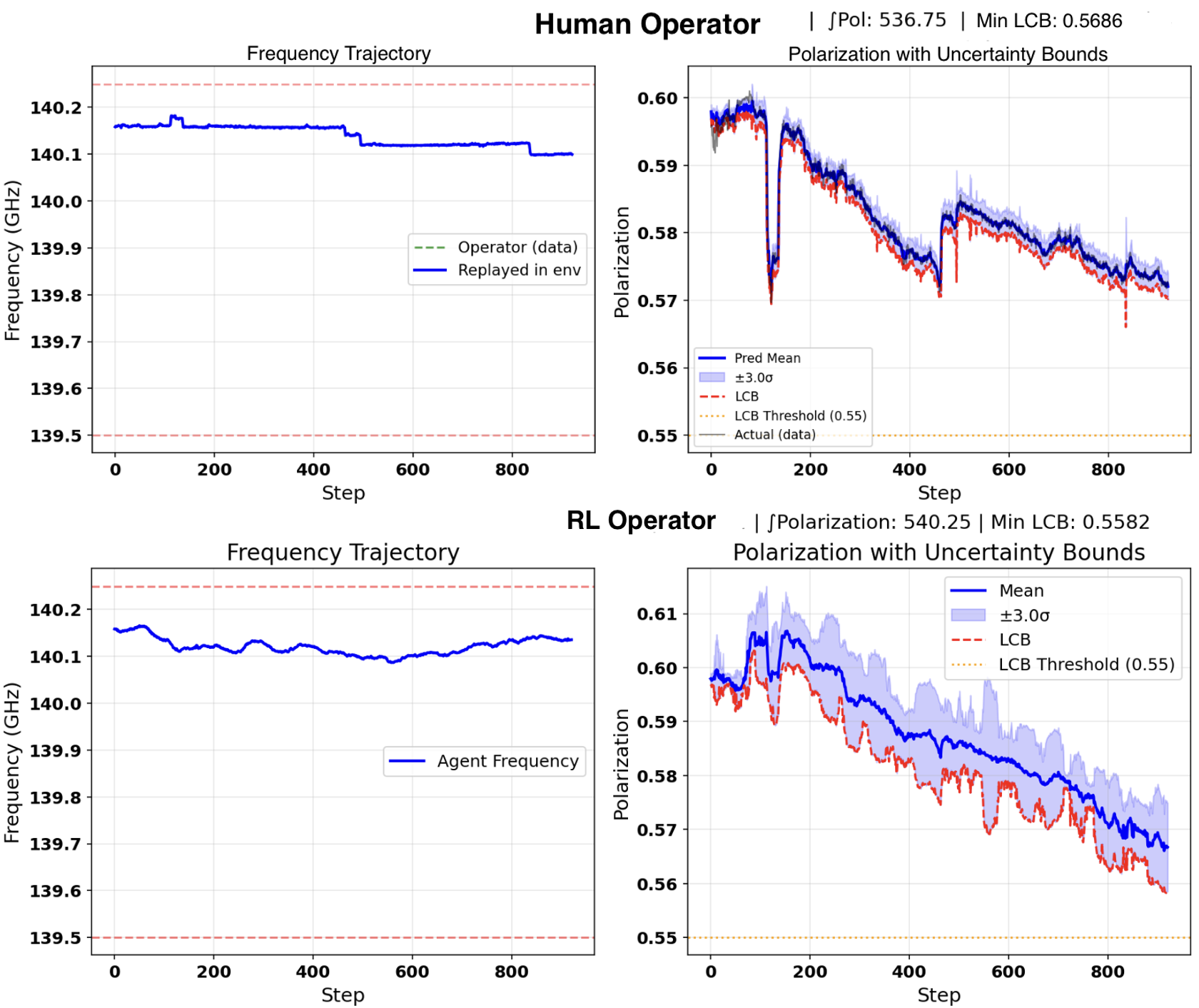}
    \caption{Human operator vs RL learned policy. Human operators actions are matched from the dataset while the RL agent is generating the actions through the policy model. These action are passed through the environment and show that that the total polarization over the episode is about equivalent}
    \label{fig:SSGPARL_VS_HUMAN}
\end{figure}

Initial training was conducted using the environment configured with the single sample GP model with the Laplace kernel, trained exclusively on data from run period P07. As shown in Figure~\ref{fig:SSGPARL_VS_HUMAN}, the trained RL agent achieved performance comparable to the human operator baseline, with no significant improvement in cumulative polarization or exploration of alternative frequency trajectories.

This result is a direct consequence of the LCB reward function (Equation~\ref{eq:LCBNORM}), which penalizes exploration into regions of high model uncertainty. Since the single sample GP is only confident near states observed during human operation, the agent is inherently constrained to similar trajectories.

\subsection{Multi Sample Model}

To address the limited exploration observed with the single sample GP, we utilized a multi sample Gaussian Process Approximation (GPA) trained on data from run periods P07, P10, and P14. The multi sample GPA generalizes the polarization state space more broadly as seen in Figure~\ref{fig:prediction_interval_GPA}, but introduces a significant computational bottleneck. As shown in Table~\ref{tab:gpa_timing}, single-step prediction time increases by over $33\times$ due to the larger training dataset and more complex covariance matrix inversions, making direct GPA integration intractable for RL training which typically requires hundreds of thousands to millions of environment interactions.

Two approaches were taken to address this runtime overhead: a precomputed lookup table (LUT) and a differentiable environment.

\begin{table}[htbp]
\centering
\label{tab:gpa_timing}
\begin{tabular}{|l|c|c|}
\hline
\textbf{Method} & \textbf{Prediction Time (ms)} & \textbf{Slowdown} \\
\hline
Single Sample GP & 32.94 & 1.00$\times$ \\
Multi Sample GPA & 1103.38 & 33.49$\times$ \\
\hline
\end{tabular}
\caption{Single-Sample GP vs. Multi-Sample GPA Prediction Time Comparison}
\end{table}

Two approaches were taken to solve the runtime overhead of the GPA: The Look up table (LUT), and differentiability.

\subsubsection{Approach 1: Precomputed Lookup Tables}

To circumvent the GPA's runtime limitations while retaining its improved uncertainty estimates, we developed a precomputed lookup table (LUT) approach that exploits the finite and discretizable nature of the environment's state-action space. The discretization parameters are shown in Table~\ref{tab:lut_discretization}.

\begin{table}[htbp]
\centering
\caption{Lookup Table Discretization Parameters}
\label{tab:lut_discretization}
\begin{tabular}{|l|c|}
\hline
\textbf{Parameter} & \textbf{Value} \\
\hline
Temporal Steps & 950 \\
Frequency Range & [139.50, 140.25] GHz \\
Frequency Granularity & 0.0001 GHz \\
Frequency States per Step & 7,500 \\
Total State Space & 7,125,000 \\
\hline
\end{tabular}
\end{table}

The key insight enabling this approach is that, unlike RL training where we must follow the agent's trajectory sequentially, LUT generation allows all 7,500 frequency states per time step to be batch processed through a single covariance matrix inversion. Using this parallelization, the complete LUT containing mean and standard deviation predictions for all 7,125,000 states was generated in approximately 30 hours on an Apple M2 Max processor. Since the LUT uses the same minimum frequency granularity as the GPA, the two produce numerically identical results. The complete state space is stored in a 43.2 MB numpy file.

\begin{figure}[htbp]
    \centering
    \includegraphics[width=1.0\linewidth]{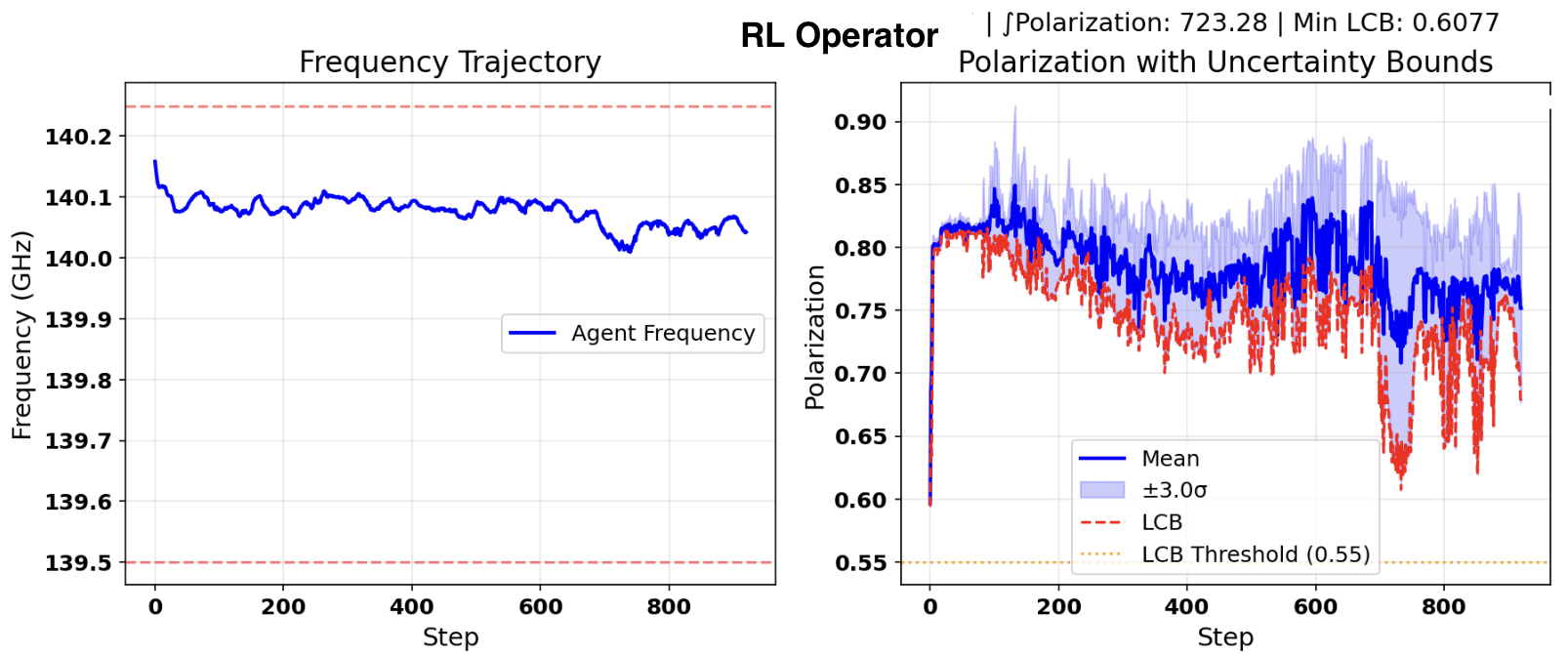}
    \caption{Multi sample GPA RL learned policy.}
    \label{fig:MSGPARL}
\end{figure}

With the LUT integrated into the environment, prediction time per step dropped to effectively zero. As shown in Figure~\ref{fig:MSGPARL}, the RL agent trained on the multi sample LUT environment found a solution achieving 33\% higher cumulative polarization than both the human operator and the single sample trained agent (Figure~\ref{fig:SSGPARL_VS_HUMAN}). These results validate our hypothesis that reducing model uncertainty through multi sample GPAs enables more effective RL exploration, allowing the agent to discover more optimal frequency trajectories.

\subsubsection{Approach 2: Differentiable Environment}

An alternative approach to addressing the GPA's computational cost is to reduce the number of environment interactions required. Gradients from the environment can aid learning in complex state spaces as seen in previous work done in high dimensional state and action spaces \cite{RLMOO_Rajput_2025} by improving sample efficiency. If gradients flow directly from the environment into the actor's learning process, we can decrease the samples required to learn a valid policy and reduce total training time on the multi sample GPA.

In standard actor-critic training (Algorithm~\ref{alg:actor_training_critic}), the actor learns by backpropagating through a frozen critic network that estimates state-action values. The critic serves as a proxy for the environment, approximating long-term expected rewards via the Bellman equation.

\begin{algorithm}
\caption{Critic-Based Actor Training Loop}
\label{alg:actor_training_critic}
\begin{algorithmic}[1]
\STATE Agent receives batch of states $\mathbf{S} = \{s_1, s_2, \ldots, s_n\}$
\STATE \textbf{Begin gradient tracking}
\STATE Pass states through actor model: $\mathbf{A} = \pi_\theta(\mathbf{S})$
\STATE Pass state-action pairs through critic model (frozen):
\STATE \hspace{1cm} $Q_\phi(\mathbf{S}, \mathbf{A})$ where $\phi$ is fixed
\STATE Compute loss from Q-values: $\mathcal{L} = -\frac{1}{n}\sum_{i=1}^{n} Q_\phi(s_i, a_i)$
\STATE Backpropagate through actor only (critic weights frozen)
\STATE Compute gradients: $\nabla_\theta \mathcal{L}$
\STATE Update actor weights: $\theta \leftarrow \theta - \alpha \nabla_\theta \mathcal{L}$
\end{algorithmic}
\end{algorithm}

We explore two strategies for incorporating environment gradients into the learning process. The first, Differentiable Trajectory Optimization (Algorithm~\ref{alg:actor_training_gradient}), eliminates the critic entirely and optimizes the actor through multi-step differentiable rollouts using only environment gradients. The second, Differentiable TD3 (Algorithm~\ref{alg:diff_td3}), retains the full TD3 framework but augments it with multi-step differentiable rollouts through the environment, combining model-based gradient information with critic-based long-horizon value estimation.

Both approaches require two key modifications to the environment. First, all environment code must be converted to support gradient tracking through TensorFlow-compatible operations, removing conditionals and casts that would break the computation graph. Second, the environment's step function must be redesigned to accept both a batch of actions and their corresponding states as input, enabling stateless forward passes. This is necessary because gradients must flow backward through the entire computation from reward, through the GP(A) dynamics, to the actor's parameters. This requires the environment to be part of the computation graph rather than maintaining decoupled internal state.

\paragraph{Differentiable Trajectory Optimization.}

The simplest use of environment gradients replaces the critic with the differentiable environment itself (Algorithm~\ref{alg:actor_training_gradient}). Rather than estimating long-term value through the Bellman equation, the actor optimizes discounted rewards accumulated over an $H$-step differentiable rollout through the environment. At each step, the actor selects an action, the environment produces the next state and reward via a differentiable forward pass, and the discounted reward is accumulated into the loss. The GP(A) model provides the gradient pathway from rewards through the multi-step dynamics back to the actor's parameters, replacing the role of the critic with direct model-based gradient descent. The rollout horizon $H$ is a hyperparameter controlling the depth of lookahead available to the actor.

\begin{algorithm}
\caption{Differentiable Trajectory Optimization}
\label{alg:actor_training_gradient}
\begin{algorithmic}[1]
\STATE Agent receives batch of states $\mathbf{S} = \{s_1, s_2, \ldots, s_n\}$ and rollout horizon $H$
\STATE \textbf{Begin gradient tracking}
\STATE $\mathcal{L} \leftarrow 0$, \quad $\gamma_{\text{pow}} \leftarrow 1$, \quad $s_{\text{cur}} \leftarrow \mathbf{S}$
\FOR{$t = 0$ to $H - 1$}
    \STATE Pass states through actor model: $\mathbf{A}_t = \pi_\theta(s_{\text{cur}})$
    \STATE Environment performs differentiable forward pass: $s_{\text{cur}},\; \mathbf{R}_t \leftarrow f(s_{\text{cur}}, \mathbf{A}_t)$
    \STATE $\mathcal{L} \leftarrow \mathcal{L} - \gamma_{\text{pow}} \cdot \frac{1}{n}\sum_{i=1}^{n} r_{t,i}$
    \STATE $\gamma_{\text{pow}} \leftarrow \gamma_{\text{pow}} \cdot \gamma$
\ENDFOR
\STATE Backpropagate through entire $H$-step computation graph (environment + actor)
\STATE Compute gradients: $\nabla_\theta \mathcal{L}$
\STATE Update actor weights: $\theta \leftarrow \theta - \alpha \nabla_\theta \mathcal{L}$
\end{algorithmic}
\end{algorithm}

This approach transforms the problem from temporal difference learning into direct trajectory optimization over a finite horizon. With multi-step rollouts, the actor can observe how its actions propagate through the environment dynamics over several time steps, partially addressing the nearsightedness of a single-step formulation. In practice, this approach converges to the optimal solution in fewer episode steps than the standard RL approach. However, it requires more wall-clock time, as each training update involves $H$ sequential environment steps through the differentiable GP(A) model. Additionally, the rollout steps taken during training occur within the optimization loop rather than the episode loop, meaning they are not counted toward the agent's episode step budget. This distinction is arguably a point of contention: the environment is still being queried $H$ times per training update, regardless of whether those interactions are logged as episode steps.

\paragraph{Differentiable TD3.}

To recover long-horizon reasoning while retaining the benefits of environment gradients, we introduce Differentiable TD3 (Algorithm~\ref{alg:diff_td3}). This approach augments the standard TD3 framework with multi-step differentiable rollouts through the environment, rather than replacing the critic entirely. The key insight is that the differentiable environment and the critic serve complementary roles: the environment provides exact local gradient information over a finite rollout horizon $H$, while the critic estimates value beyond that horizon.

In the actor update, rather than backpropagating through a frozen critic alone (as in standard TD3), the actor unrolls $H$ steps through the differentiable environment, accumulating discounted rewards at each step. At the end of the rollout, the critic provides a bootstrap value estimate for the remaining infinite horizon. Gradients flow backward through the entire $H$-step chain, from the critic bootstrap through each environment step back to the actor's parameters, giving the actor direct access to how its actions affect future states and rewards over multiple time steps.

The environment's differentiable step function also returns a soft termination signal, $p_{\text{alive}}$, implemented as a sigmoid over the distance from the termination boundary (the lower confidence bound threshold). This signal multiplicatively attenuates future rewards as the trajectory approaches termination:
\begin{equation}
    \bar{p}_t = \prod_{k=1}^{t} p_{\text{alive},k}
\end{equation}
Because the sigmoid is differentiable, the actor receives gradient signal indicating that actions leading toward low-confidence regions will diminish all downstream rewards, creating a smooth incentive to avoid the termination boundary without discontinuous masking.

The critic update mirrors this structure. Rather than computing single-step Bellman targets from replay buffer transitions, the target value is constructed by rolling out $H$ steps from the sampled next state using the target policy, accumulating discounted rewards attenuated by the survival probability, and bootstrapping at the terminal state with the clipped double-Q target:
\begin{equation}
\resizebox{0.95\columnwidth}{!}{$\displaystyle
    y = r_0 + \sum_{t=1}^{H-1} \gamma^t\, r_t\, \bar{p}_t \;+\; \gamma^H \min\bigl(Q_{\phi_1'}(s_H, \tilde{a}_H),\; Q_{\phi_2'}(s_H, \tilde{a}_H)\bigr)\, \bar{p}_{H-1}
$}
\end{equation}
where $\tilde{a}_H = \pi_{\theta'}(s_H) + \epsilon$ is the smoothed target action. This provides the critic with richer multi-step target signals grounded in the true environment dynamics, rather than relying solely on single-step bootstrapping.

\begin{algorithm}
\caption{Differentiable TD3 (Diff-TD3)}
\label{alg:diff_td3}
\begin{algorithmic}[1]
\STATE Initialize actor $\pi_\theta$, critics $Q_{\phi_1}, Q_{\phi_2}$, target networks $\pi_{\theta'}, Q_{\phi_1'}, Q_{\phi_2'}$, replay buffer $\mathcal{B}$, rollout horizon $H$
\FOR{each training step}
    \STATE Sample batch $(s, a, r_0, s', d)$ from $\mathcal{B}$
    \STATE
    \STATE \textit{// Critic update with H-step model-based targets }
    \STATE $\bar{p} \leftarrow (1 - d)$, \quad $y \leftarrow r_0$, \quad $s_{\text{cur}} \leftarrow s'$
    \FOR{$t = 1$ to $H - 1$}
        \STATE $\tilde{a}_t \leftarrow \text{clip}\bigl(\pi_{\theta'}(s_{\text{cur}}) + \text{clip}(\epsilon, -c, c),\; a_{\text{low}},\; a_{\text{high}}\bigr)$
        \STATE $s_{\text{cur}},\; r_t,\; p_{\text{alive}} \leftarrow f'(s_{\text{cur}},\; \tilde{a}_t)$ \hfill \textit{// differentiable env step}
        \STATE $\bar{p} \leftarrow \bar{p} \cdot p_{\text{alive}}$
        \STATE $y \leftarrow y + \gamma^t\, r_t\, \bar{p}$
    \ENDFOR
    \STATE $\tilde{a}_H \leftarrow \text{clip}\bigl(\pi_{\theta'}(s_{\text{cur}}) + \text{clip}(\epsilon, -c, c),\; a_{\text{low}},\; a_{\text{high}}\bigr)$
    \STATE $y \leftarrow y + \gamma^H \min\bigl(Q_{\phi_1'}(s_{\text{cur}}, \tilde{a}_H),\; Q_{\phi_2'}(s_{\text{cur}}, \tilde{a}_H)\bigr) \cdot \bar{p}$
    \STATE Update $\phi_1, \phi_2$ by minimizing $\frac{1}{n}\sum_i \bigl(Q_{\phi_j}(s_i, a_i) - y_i\bigr)^2$ for $j = 1, 2$
    \STATE
    \STATE \textit{// Actor update with H-step differentiable rollout}
    \IF{$t \mod d_{\text{actor}} = 0$}
        \STATE \textbf{Begin gradient tracking}
        \STATE $J \leftarrow 0$, \quad $\bar{p} \leftarrow 1$, \quad $s_{\text{cur}} \leftarrow s$
        \FOR{$t = 0$ to $H - 1$}
            \STATE $a_t \leftarrow \pi_\theta(s_{\text{cur}})$
            \STATE $s_{\text{cur}},\; r_t,\; p_{\text{alive}} \leftarrow f'(s_{\text{cur}},\; a_t)$ \hfill \textit{// differentiable env step}
            \STATE $\bar{p} \leftarrow \bar{p} \cdot p_{\text{alive}}$
            \STATE $J \leftarrow J + \gamma^t\, r_t\, \bar{p}$
        \ENDFOR
        \STATE $J \leftarrow J + \gamma^H\, Q_{\phi_1}(s_{\text{cur}},\; \pi_\theta(s_{\text{cur}}))\, \bar{p}$ \hfill \textit{// critic bootstrap}
        \STATE Update $\theta$ by maximizing $J$ (gradients flow through env + critic)
        \STATE Soft-update target networks: $\theta' \leftarrow \tau\theta + (1-\tau)\theta'$, etc.
    \ENDIF
\ENDFOR
\end{algorithmic}
\end{algorithm}

The critical distinction between Differentiable TD3 and Differentiable Trajectory Optimization lies in how they extend the actor's planning horizon. Both perform $H$-step differentiable rollouts through the environment, but Differentiable TD3 additionally bootstraps with a learned critic at step $H$ to approximate value beyond the rollout horizon, while Trajectory Optimization relies entirely on the accumulated rollout rewards.

In practice on this environment, Differentiable TD3 converges in approximately the same number of episode steps as standard TD3, while incurring greater wall-clock time due to the additional cost of unrolling $H$ differentiable environment steps per training update. We attribute this to two factors. First, the GP(A) environment is sufficiently simple that the critic alone provides adequate long-term value estimation, i.e. the additional gradient signal from the differentiable rollout does not meaningfully accelerate convergence in step count. Second, the state-action space is not high-dimensional enough to require environment gradients for efficient exploration, unlike the settings described in \cite{RLMOO_Rajput_2025} where differentiable gradients provided substantial benefit. Differentiable Trajectory Optimization, by contrast, converges in fewer episode steps than standard TD3 but at greater wall-clock cost, suggesting that the direct environment gradients do improve per-step learning efficiency. These results indicate that the relative advantage of each approach is likely environment-dependent: in higher-dimensional or more complex settings where the critic struggles to approximate the value landscape accurately, the augmented gradient information from Differentiable TD3 may yield more substantial improvements.

\section{7 Conclusion}

We've shown that the utilization of data driven surrogate models with uncertainties can be leveraged to create environments where we can train RL agents to learn better control of the microwave frequency than the historical actions taken from an operator during the P07 run period. Reinforcement learning has been proven to make effective control decisions for polarization maximization challenges found in the control room. Future research will look into solving the challenges found in operational integration and preparation as well as more effective RL algorithms that utilize both the Bellman equations discounted long term rewards and gradients from the environment.

\section{Acknowledgments}
\begin{acknowledgments}
This manuscript has been authored by Jefferson Science Associates (JSA) operating
the Thomas Jefferson National Accelerator Facility for the U.S. Department of Energy
under Contract No. DE-AC05-06OR23177. The authors acknowledge support by the
U.S. Department of Energy, Office of Science. The US government retains and the
publisher, by accepting the article for publication, acknowledges that the US government
retains a nonexclusive, paid-up, irrevocable, worldwide license to publish or reproduce
the published form of this manuscript, or allow others to do so, for US government
purposes. DOE will provide public access to these results of federally sponsored research
in accordance with the DOE Public Access Plan (http://energy.gov/downloads/doe-
public-access-plan).
\end{acknowledgments}

\bibliographystyle{apsrev4-1}
\bibliography{references.bib}

@book{sutton_barto_rl2018,
author = {Sutton, Richard S. and Barto, Andrew G.},
title = {Reinforcement Learning: An Introduction},
year = {2018},
isbn = {978-0262039246},
publisher = {MIT Press},
address = {Cambridge, MA, USA},
edition   = 2,
}

@unpublished{relu,
  author = {Agarap, Abien Fred},
  title = "Deep Learning using Rectified Linear Units ({ReLU})",
  eprint = {1803.08375},
  archivePrefix = "arXiv",
  year = {2018}
}

@article{chung2021uncertainty,
  title={Uncertainty Toolbox: an Open-Source Library for Assessing, Visualizing, and Improving Uncertainty Quantification},
  author={Chung, Youngseog and Char, Ian and Guo, Han and Schneider, Jeff and Neiswanger, Willie},
  journal={arXiv preprint arXiv:2109.10254},
  year={2021}
}

@misc{rl_td3,
  doi = {10.48550/ARXIV.1802.09477},
  
  url = {https://arxiv.org/abs/1802.09477},
  
  author = {Fujimoto, Scott and van Hoof, Herke and Meger, David},
  
  title = {Addressing Function Approximation Error in Actor-Critic Methods},
  
  publisher = {arXiv},
  
  year = {2018},
  
  copyright = {arXiv.org perpetual, non-exclusive license}
}

@article{RevModPhys.94.031003,
  title = {Colloquium: Machine learning in nuclear physics},
  author = {Boehnlein, Amber and Diefenthaler, Markus and Sato, Nobuo and Schram, Malachi and Ziegler, Veronique and Fanelli, Cristiano and Hjorth-Jensen, Morten and Horn, Tanja and Kuchera, Michelle P. and Lee, Dean and Nazarewicz, Witold and Ostroumov, Peter and Orginos, Kostas and Poon, Alan and Wang, Xin-Nian and Scheinker, Alexander and Smith, Michael S. and Pang, Long-Gang},
  journal = {Rev. Mod. Phys.},
  volume = {94},
  issue = {3},
  pages = {031003},
  numpages = {32},
  year = {2022},
  month = {Sep},
  publisher = {American Physical Society},
  doi = {10.1103/RevModPhys.94.031003},
  url = {https://link.aps.org/doi/10.1103/RevModPhys.94.031003}
}

@article{GOERTZ2002403,
title = {Polarized H, D and 3He targets for particle physics experiments},
journal = {Progress in Particle and Nuclear Physics},
volume = {49},
number = {2},
pages = {403-489},
year = {2002},
issn = {0146-6410},
doi = {https://doi.org/10.1016/S0146-6410(02)00159-X},
url = {https://www.sciencedirect.com/science/article/pii/S014664100200159X},
author = {St. Goertz and W. Meyer and G. Reicherz}
}

@article{Pandey:2024Wu,
  author = "Pandey, Pushpa  and  Brock, James  and  Kageya, Tsuneo  and  Keith, Christopher  and  Kuhn, Sebastian E  and  Lagerquist, Victoria  and  Maxwell, James  and  Wei, Xiangdong",
  title = "{Operation of a Longitudinally Polarized Solid Nuclear Target in CLAS12}",
  doi = "10.22323/1.456.0211",
  journal = "PoS",
  year = 2024,
  volume = "SPIN2023",
  pages = "211"
}

@article{maxwell_2026,
	title = {A new cw-{NMR} {Q}-meter for dynamically polarized targets for particle physics},
	volume = {1087},
	issn = {0168-9002},
	url = {https://www.sciencedirect.com/science/article/pii/S0168900226001439},
	doi = {10.1016/j.nima.2026.171417},
	urldate = {2026-03-09},
	journal = {Nuclear Instruments and Methods in Physics Research Section A: Accelerators, Spectrometers, Detectors and Associated Equipment},
	author = {Maxwell, J. D. and Brock, J. and Cuevas, C. and Dong, H. and Keith, C. D. and Pierce, J. J.},
	month = jul,
	year = {2026},
	pages = {171417}
}

@misc{towers2025gymnasiumstandardinterfacereinforcement,
      title={Gymnasium: A Standard Interface for Reinforcement Learning Environments}, 
      author={Mark Towers and Ariel Kwiatkowski and Jordan Terry and John U. Balis and Gianluca De Cola and Tristan Deleu and Manuel Goulão and Andreas Kallinteris and Markus Krimmel and Arjun KG and Rodrigo Perez-Vicente and Andrea Pierré and Sander Schulhoff and Jun Jet Tai and Hannah Tan and Omar G. Younis},
      year={2025},
      eprint={2407.17032},
      archivePrefix={arXiv},
      primaryClass={cs.LG},
      url={https://arxiv.org/abs/2407.17032}, 
}

@article{sac,
  author       = {Tuomas Haarnoja and
                  Aurick Zhou and
                  Pieter Abbeel and
                  Sergey Levine},
  title        = {Soft Actor-Critic: Off-Policy Maximum Entropy Deep Reinforcement Learning
                  with a Stochastic Actor},
  journal      = {CoRR},
  volume       = {abs/1801.01290},
  year         = {2018},
  url          = {http://arxiv.org/abs/1801.01290},
  eprinttype    = {arXiv},
  eprint       = {1801.01290},
  bibsource    = {dblp computer science bibliography, https://dblp.org}
}

@article{td3,
  author       = {Scott Fujimoto and
                  Herke van Hoof and
                  David Meger},
  title        = {Addressing Function Approximation Error in Actor-Critic Methods},
  journal      = {CoRR},
  volume       = {abs/1802.09477},
  year         = {2018},
  url          = {http://arxiv.org/abs/1802.09477},
  eprinttype    = {arXiv},
  eprint       = {1802.09477},
  bibsource    = {dblp computer science bibliography, https://dblp.org}
}

@inproceedings{Silver2014DeterministicPG,
  title={Deterministic Policy Gradient Algorithms},
  author={David Silver and Guy Lever and Nicolas Manfred Otto Heess and Thomas Degris and Daan Wierstra and Martin A. Riedmiller},
  booktitle={International Conference on Machine Learning},
  year={2014},
  url={https://api.semanticscholar.org/CorpusID:13928442}
}

@article{RLMOO_Rajput_2025,
doi = {10.1088/2632-2153/adc221},
url = {https://doi.org/10.1088/2632-2153/adc221},
year = {2025},
month = {apr},
publisher = {IOP Publishing},
volume = {6},
number = {2},
pages = {025018},
author = {Rajput, Kishansingh and Schram, Malachi and Edelen, Auralee and Colen, Jonathan and Kasparian, Armen and Roussel, Ryan and Carpenter, Adam and Zhang, He and Benesch, Jay},
title = {Harnessing the power of gradient-based simulations for multi-objective optimization in particle accelerators},
journal = {Machine Learning: Science and Technology}
}

@inproceedings{kendall2017uncertainties,
  title={What Uncertainties Do We Need in Bayesian Deep Learning for Computer Vision?},
  author={Kendall, Alex and Gal, Yarin},
  booktitle={Proceedings of the 31st Conference on Neural Information Processing Systems (NeurIPS 2017)},
  year={2017},
  url={https://arxiv.org/abs/1703.04977},
  note={arXiv:1703.04977v2 [cs.CV]}
}

@article{chen2024uncertainty,
  title        = {Uncertainty-Aware Out-of-Distribution Detection with Gaussian Processes},
  author       = {Chen, Yang and Sung, Chih-Li and Kusari, Arpan and Song, Xiaoyang and Sun, Wenbo},
  journal      = {arXiv preprint arXiv:2412.20918},
  year         = {2024},
  url          = {https://arxiv.org/abs/2412.20918}
}

@article{schram2023uncertainty,
  title = {Uncertainty aware machine-learning-based surrogate models for particle accelerators: Study at the Fermilab Booster Accelerator Complex},
  author = {Schram, Malachi and Rajput, Kishansingh and Somayaji, Karthik S. and Li, Peng and St. John, Jason and Sharma, Himanshu},
  journal = {Physical Review Accelerators and Beams},
  volume = {26},
  number = {4},
  pages = {044602},
  year = {2023},
  doi = {10.1103/PhysRevAccelBeams.26.044602}
}

@book{rasmussen2006gaussian,
  title     = {Gaussian Processes for Machine Learning},
  author    = {Rasmussen, Carl Edward and Williams, Christopher K.~I.},
  year      = {2006},
  publisher = {MIT Press},
  address   = {Cambridge, MA, USA},
  url       = {https://gaussianprocess.org/gpml/chapters/RW.pdf}
}

@article{dai2020federated,
  title        = {Federated Bayesian Optimization via Thompson Sampling},
  author       = {Dai, Zhongxiang and Low, Kian Hsiang and Jaillet, Patrick},
  journal      = {arXiv preprint},
  volume       = {arXiv:2010.10154},
  year         = {2020},
  url          = {https://arxiv.org/abs/2010.10154},
  eprint       = {2010.10154},
  archivePrefix= {arXiv},
  primaryClass = {cs.LG}
}

@InProceedings{snelson2007local,
  title     = {Local and Global Sparse Gaussian Process Approximations},
  author    = {Snelson, Edward and Ghahramani, Zoubin},
  booktitle = {Proceedings of the Eleventh International Conference on Artificial Intelligence and Statistics},
  series    = {Proceedings of Machine Learning Research},
  volume    = {2},
  pages     = {524--531},
  year      = {2007},
  editor    = {Meila, Marina and Shen, Xiaotong},
  address   = {San Juan, Puerto Rico},
  publisher = {PMLR},
  url       = {https://proceedings.mlr.press/v2/snelson07a.html}
}

\clearpage
\newpage
\newpage

\setcounter{figure}{0}
\renewcommand{\thefigure}{S\arabic{figure}}

\setcounter{table}{0}
\renewcommand{\thetable}{S\arabic{table}}

\setcounter{equation}{0}
\renewcommand{\theequation}{S\arabic{equation}}

\end{document}